\documentclass[final]{cvpr}
\usepackage{times}
\usepackage{epsfig}
\usepackage{graphicx}
\usepackage{amsmath}
\usepackage{amssymb}
\usepackage[pagebackref=true,breaklinks=true,colorlinks,bookmarks=false]{hyperref}
\usepackage[ruled,vlined,linesnumbered]{algorithm2e}
\usepackage[noend]{algpseudocode}
\usepackage{amsmath,amssymb,amsthm,float,,caption,mathtools,relsize}
\usepackage{enumerate}
\usepackage{subfig}
\usepackage{bold-extra}
\usepackage{caption}
\usepackage{paralist}
\usepackage{mathtools}
\usepackage{multirow}
\usepackage{layouts}

\def\cvprPaperID{****} 
\def\confYear{CVPR 2021}
\begin{document}
%textwidth in cm: \printinunitsof{in}\prntlen{\textwidth}
%%%%%%%%% TITLE
\title{Efficacy of Bayesian Neural Networks in Active Learning}

\author{Vineeth Rakesh\\
Interdigital AI Lab\\
USA\\
{\tt\small vineethrakesh@gmail.com}
\and
Swayambhoo Jain\\
Interdigital AI Lab\\
USA\\
{\tt\small swayambhoo.jain@gmail.com}
}

\maketitle

\begin{abstract}
Obtaining labeled data for machine learning tasks can be prohibitively expensive. Active learning mitigates this issue by exploring the unlabeled data space and prioritizing the selection of data that can best improve the model performance. A common approach to active learning is to pick a small sample of data for which the model is most \textit{uncertain}. In this paper, we explore the efficacy of Bayesian neural networks for active learning, which naturally models uncertainty by learning distribution over the weights of neural networks. By performing a comprehensive set of experiments, we show that Bayesian neural networks are more efficient than ensemble based techniques in capturing uncertainty. Our findings also reveal some key drawbacks of the ensemble techniques, which was recently shown to be more effective than Monte Carlo dropouts.
\end{abstract}

%With just $10\%$ of data-points for Fashion MNIST and with just $40\%$ for Cifar10, Bayesian neural networks can achieve approximately the same accuracy as a model trained over the entire dataset. However, estimating uncertainty in Bayesian networks are computationally expensive. Therefore, we provide an efficient uncertainty approximation algorithm for Bayesian neural networks with dense layers which requires just a single-pass through the network. We show that our algorithm is capable of achieving performances that is on-par with the traditional uncertainty estimation in active learning. both in terms of overall accuracy and class-specific . continual training and on-par with ensemble, which needs atleast four to five times the number of parameters. which results in significant gains in active learning performance.
%when retrained in a continual fashion and are on-par with ensembles. With the trade-off of just doubling the number of parameters, BNNs ...over ensemble, which requires atleast four to five individual neural networks to be trained. 

\section{Introduction}

Although machine learning techniques have achieved a major breakthrough in recent years, their performance comes at a cost of acquiring large volumes of training data. This is especially true for supervised deep learning models that demand a substantial amount of labeled data to achieve a reasonable performance. For applications that require expert knowledge such as medical and biological images, labels are extremely hard and expensive to obtain. Active learning (AL) aims to mitigate this problem by smartly selecting data points to label (from an expert) from a large pool of unlabeled data to improve model performance. This sampling is typically based on some \textit{acquisition function} (AF) which provides a score for each unlabeled data that signifies its level of importance. While there are many approaches to implementing AF \cite{sener2017geometric,yang2017suggestive},  uncertainty-based based approaches are shown to be the most effective \cite{ yang2015multi, lakshminarayanan2017simple, gal2017deep, beluch2018power}. 

Bayesian neural network (BNN) naturally models uncertainty by learning a probability distribution over the neural network weights. Therefore, for a given input, as we take multiple realizations of the network, the variance captured by the weights is reflected as the variation in the output, which in-turn models uncertainty. BNNs learn by applying a prior distribution over weights and performing variational inference to approximate the posterior distribution. In \cite{gal2016dropout}, the authors proved that applying dropout to neural networks is equivalent to a BNN. This theory was further leveraged by proposing Monte Carlo Dropout (MCD) for uncertainty estimation in AL  \cite{gal2017deep}. In a recent work,  \cite{beluch2018power} showed that ensemble of neural networks (EN) outperform MCD when it comes to uncertainty estimation; thus, proving to be the choice for active learning. Consequentially, it is natural to assume EN to perform better than BNN since MCD is equivalent to BNN.  \textit{However, dropout neural networks form a special class of BNN where the posterior distribution is a special case of \emph{spike-and-slab} distribution. Contrary to this, BNNs allow for a broader class of prior and posterior distributions on weights}.

% While BNNs are theoretically well motivated for uncertainty estimation
In this paper, we re-establish the efficacy of BNNs in active learning over ensembles and MCD by using a more general scaled normal prior based BNN proposed in \cite{louizos2017bayesian}. The scaled normal prior is a continuous relaxation of the \emph{spike-and-slab} distribution and subsumes Dropout as a special case.    Through extensive  experiments on multiple datasets namely, MNIST, Fashion MNIST, CIFAR10 and CIFAR100 and a regression dataset on housing price prediction we show that the scaled normal prior based BNN provides more robust and efficient active learning over EN and MCD.  We perform several experiments to demonstrate the pros and cons of BNN over EN and MCD. For each round of active learning, the models are trained using two different settings: 
\begin{inparaenum}[(1)]
	\item re-use the trained state of the model from previous round and retrain on the newly appended datapoints (termed as \textit{continual training}) and
	\item reset the model parameters and retrain from scratch.
\end{inparaenum} Our results show that BNN performs significantly better than EN in terms of classification accuracy when it comes to continual training. In fact, the performance of EN is worse than MCD which can be attributed to overfitting and catastrophic forgetting. That being said, when retrained from scratch, BNN and EN perform on a similar level, which is still an advantage for BNN since estimating uncertainty using ensembles is a costly process. We found that EN requires about five ensembles in order to achieve good active learning performance.  This implies, training of five different i.i.d networks and storing the trained state of every single network instance. BNN on other hand, achieves similar yet a more robust performance with a trade-off of just doubling parameter size of conventional neural network. 

Besides illustrating the overall effectiveness of BNN for active learning, we answer the following questions:
\begin{inparaenum}[(1)]
	\item do acquisition functions behave the same for Bayesian, ensemble and MC dropouts?
	\item how does model capacity affect the outcome, do BNNs with lower model capacity perform worse than EN (or MCD)?
	\item are BNNs better than EN when predicting challenging class labels?
\end{inparaenum}
Inspired by the performance of BNNs, we also propose a computationally efficient uncertainty estimation method for fully connected dense layers with ReLU non-linearity. Since AL involves repeated uncertainty estimation over large unlabeled dataset, efficient uncertainty estimation is of huge practical importance. In the proposed method, instead of taking multiple instantiations of neural networks to estimate the uncertainty, we perform just one forward pass. In this forward pass, at each neuron, we approximate the probability distribution parametrically. We show that this algorithm is capable of achieving performances that is on-par with the traditional uncertainty estimation in BNN.

To the best of our knowledge, we are the first to perform comprehensive empirical analysis to demonstrate the efficacy of BNNs for active learning. While most existing research limit themselves to experiments on small architectures and dataset, ours does not have such constraints. 

%We implement Bayesian versions of large neural network and evaluates on the most popular image classification dataset. We implement the Bayesian versions of the point estimate neural networks, and compare them with their EN and MCD counterparts. 

%comparison between BNNs and MCD for active learning. Although there are some works \cite{beluch2018power} that briefly discuss the use of BNNs for active learning, there is still a lack of detailed comparison between these two techniques and in this paper our objective is to fill this void.
%We believe our experiments can aid researchers and practitioners to understand the scope of BNNs in active learning. 
%large scale

%. (b) and (c) the empirical experiments are extremely basic. We believe our rigorous ...will help researchers and practitioners to get a deeper understanding of BNNs for active learning, show their strengths and drawbacks. 
%These studies either propose a scalable algorithm for active learning and their focus is not primarily on demonstrating the efficacy of BNNs over Ensemble or Mc Dropouts

%

%To validate the efficacy of BNNs, we perform different experiments starting from simple dense layers to deeper convolution neural networks (CNN) such as VGG and Densenet. 

% Unfortunately, BCN suffers from poor calibration (e)-(h), but it can be quite easily corrected. In particular,  (b) Bayes closed-form network (BCN) a more efficient BNN that replaces sampling from posterior using deterministic function of variational parameters \cite{molchanov2017variational, louizos2017bayesian}. 

\section{Related Work}

%\subsection{Active Learning}
Active learning has been explored extensively in classical machine learning literature \cite{settles2009active}. Much of the focus of classical literature has been on the high dimensional data arising in the context of linear models such as support vector machine \cite{holub2008entropy, joshi2009multi, li2013adaptive}. That being said, recently, there has been a significant interest in AL for deep neural networks (DNNs). While there are many ways to perform AL in DNNs, uncertainty-based sampling techniques are the preferred choice due to their ease of implementation and computational efficiency. 
Uncertainty in the output of neural networks can be estimated using (a) Bayesian neural networks or (b) ensemble of neural networks.

% Dropout based uncertainty estimates
A popular technique for AL using BNNs is called Monte Carlo dropout which was first proposed in \cite{gal2017deep}. The basic idea is to pass the new unlabeled data through the DNNs multiple times while retaining the dropout layer. This results in different realizations at the output of the DNNs. These realizations can be used to estimate the uncertainty in output using various measures such as entropy or the variance ratio. The estimated uncertainty can then be utilized for acquiring new unlabeled data points thus performing active learning.  

On the other hand, ensemble based uncertainty estimation involves having an ensemble of neural networks which typically have same architecture but trained with different random initialization \cite{lakshminarayanan2017simple}. Uncertainty is estimated by passing unlabeled examples through individual ensembles and their outputs are then used to estimate the uncertainty. It was shown in \cite{beluch2018power} that ensemble methods outperforms the Monte Carlo dropout-based estimation.  This performance is primarily attributed to the higher capacity and diversity in the ensemble models as compared to different realizations of neural networks in dropout based networks. While there are some studies that leverage BNNs for active learning \cite{lakshminarayanan2017simple, kirsch2019batchbald,atighehchian2020bayesian,munjal2020towards}, they have a few or all of the following shortcomings:
\begin{inparaenum}[(1)]
	\item experiments are restricted to simple architectures with a few dense or convolutional neural network (CNN) layers
	\item evaluation is restricted to basic datasets such as MNIST
	\item no comparison of BNNs with ensemble models
	\item claim to use BNNs, but actually use Monte Carlo dropouts as approximation to BNNs.
\end{inparaenum}

%{\color{red} This, however, comes at the cost of high inference and training complexity in ensemble methods.  The ensemble methods require training and inference all the models which consumes memory and computation. In contrast to this, the Bayesian deep neural networks the training involves learning the parameters of the prior. Further, if prior has less number of parameters as compared to all the models in ensembles then the complexity with Bayesian deep neural network can be reduced.} 

\section{Active Learning Via Bayesian Neural Networks}
\label{sec:modeling}

{\bf Bayesian neural network}: For a given dataset $\mathcal{D} = \{(\mathbf{x}_i, \mathbf{y}_i)\}_{i=1}^{D}$, Bayesian neural networks involves calculation of the distribution of weights given the training data $p(\mathbf{w}|\mathcal{D})$. The predictive distribution for a test data $\mathbf{x}$ can be obtained by marginalizing $\mathbf{w}$ as follows: $p(\mathbf{y}|\mathbf{x}, \mathcal{D}) = \int{p(\mathbf{y}|\mathbf{x}, \mathbf{w}) p(\mathbf{w}|\mathcal{D}) d\mathbf{w}}$. This is equivalent to \textit{averaging predictions from an ensemble of neural networks} weighted by the posterior probabilities of their parameters $\mathbf{w}$. However, exact computation of the posterior is intractable, therefore, we resort to variational inference.  That is, we wish to approximate $p(\mathbf{w}|\mathcal{D}) = p(\mathcal{D}|\mathbf{w})p(\mathbf{w})/p(\mathcal{D})$ by positing an approximate posterior $q_{\phi}(\mathbf{w})$ of variational parameters $\phi$. The problem then reduces to optimizing the evidence-lower-bound (ELBO) defined as follows:
\begin{align}
\mathcal{L}(\phi) = \underbrace{\mathbb{E}_{q_{\phi}(\mathbf{w})}[\log p(\mathcal{D} | \mathbf{w})]}_{(a)} + \underbrace{\textrm{KL}[q_{\phi}(\mathbf{w}) || p(\mathbf{w})]}_{(b)}
\label{eq:elbo}
\end{align}
where term (a) is the data-dependent likelihood term, and (b) is the regularizer that measures the KL divergence between the posterior and prior.  For prior we use the scaled normal prior proposed in  \cite{louizos2017bayesian} in which the scales $z$ follow a log-uniform prior: $ p(z) \propto |z|^{-1} $. For a given layer weight matrix $\mathbf{W} \in \mathbf{R}^{m \times n}$ for a fully connected layer of neural network with input dimension $n$ and output dimension $m$
the scales of shared across input dimension as 
\begin{align}
p(\mathbf{W},\mathbf{z}) \propto \prod_{j=1}^n \frac{1}{|z_j|}  \prod_{i,j}^{m,n} \mathcal{N}(w_{ij}|0, z_j^2).
\end{align}
The main rationale behind scaled normal prior is that it is continuous relaxation of the \emph{spike-and-slab} distribution. Dropout which is the basis of MCD based active learning is a special case of \emph{spike-and-slab} distribution  \cite{louizos2017bayesian} and we hope to get better performance with the scaled normal prior. We consider the following joint approximate posterior 
\begin{align}
	q_{\phi}(\mathbf{W}, \mathbf{z}) = \prod_{j}^{n} \mathcal{N}(z_{j} | \mu_{z_j}, \sigma^{2}_{z_j}) \prod_{i,j}^{m,n} \mathcal{N}(w_{ij}|z_{j}\mu_{ij}, z_{j}^{2}\sigma_{ij}^{2}), \label{eq:BCN_Posterior}
\end{align}

and the corresponding ELBO is given by:

\begin{align}
\mathcal{L}(\phi)  &= \mathbb{E}_{q_{\phi (\mathbf{z})}q_{\phi}(\mathbf{W}|\mathbf{z})}[\text{log}\, p(\mathcal{D}|\mathbf{W})]  \label{eqn:ELBO_BCN} \\ 
- &\mathbb{E}_{q_{\phi} (\mathbf{z})}[\textrm{KL}(q_{\phi}(\mathbf{W} | \mathbf{z})) || p(\mathbf{W} | \mathbf{z})] -\textrm{KL}(q_{\phi}(\mathbf{z}) || p(\mathbf{z})).\nonumber
\end{align}
The KL divergences can be replaced with the closed form expressions
\begin{align*}
&\textrm{KL}(q_{\phi}(\mathbf{W} | \mathbf{z})) || p(\mathbf{W} | \mathbf{z}) =  \frac{1}{2} \sum_{i,j}^{m,n}  \log  \frac{e^{\sigma_{ij}^2 + \mu_{ij}^2 - 1}}{\sigma_{ij}^2}, \\
	 &\textrm{KL} (q_{\phi}(z) || p(z)) \approx \sum_{j}^{n} k_1 \left( 1 - \gamma \left(k_2 - k_3 \alpha_j \right) - \frac{ m(\alpha_j) }{2k_1} \right),
	%\label{eq:closed_kl}
\end{align*}
where  $\alpha_j = -\log(\sigma_{z_j}^2/\mu_{z_j}^2)$, $\gamma(\cdot)$ and $m(\cdot)$ are the sigmoid and soft-plus functions respectively and the constants $k_1 = 0.63576, k_2 =1.87320, k_3 = 1.48695$ \cite{molchanov2017variational}. %We refer to this Bayesian neural network with as the \underline{B}ayesian \underline{c}losed-form \underline{n}etwork (BCN).

From practical implementation point-of-view this scaled normal prior BNN is easier to implement as compared to other prior. By virtue of the closed form KL-divergences the ELBO in \eqref{eqn:ELBO_BCN} can be optimized within the framework of standard stochastic gradient ascent. In addition to this, during test time it can implemented as a single feedforward pass where we replace $\mathbf{W}$ at layer with its mean $\tilde{\mathbf{W}} = \mathbf{M}_W \textrm{diag}(\mathbf{\mu}_z)$
where $\mathbf{M}_W$ is the matrix of means $\mathbf{\mu}_{ij}$ and $\mathbf{\mu}_z$ is the vector of means $\mu_{z_j}$.

\noindent {\bf Acquisition functions}: Once the model is trained on a small dataset, we use acquisition functions (AF) to fetch the most uncertain datapoints. In a recent work \cite{beluch2018power}, the authors experimented with several acquisition functions and found entropy \cite{shannon1948mathematical} and variation-ratio to be the best candidates  \cite{freeman1965elementary}.  Therefore, we use these two metrics as our choice of AF.  A BNN trained for multi-class classification problem of $C$ classes maps a given data vector $\mathbf{x}$ to a $C$ dimensional vector containing the probability of various classes in its components. For a given unlabelled data vector $\mathbf{x}$ the entropy is calculated by $T$ forward passes, each time with new weight realization  $\mathbf{W}_t$ from the trained posterior. First, the $T$ outputs vectors are averaged to obtain the probability for a given class $c$ as $\hat{p}( y = c | \mathbf{x})  = 1/T \sum_{t} p( y = c | \mathbf{x}, \mathbf{W}_t)$  where $\mathbf{W}_t$ is the  $t^{th}$ realization of weights obtained from the trained posterior. Next, with these probability estimates the entropy can be calculated as follows:%\begin{align}
\begin{align}
	H( y | \mathbf{x} ) = - \sum_{c} \hat{p}( y=c | \mathbf{x})  \log\left(\hat{p}( y = c | \mathbf{x})\right).\label{eq:entropy}
\end{align}
The variation ratio can be calculated as $v = 1 - f_m/T$ where $f_m$ is the number of predictions falling into the modal class category.

\noindent {\bf Active learning algorithm}: With the Bayesian neural networks and acquisition functions defined, we now describe our active learning methodology in Algorithm \ref{alg:gen_prcss}. The procedure starts by training the model with some seed sample $\mathcal{S}$ (line \ref{alg:seed_sample}). The size of this sample could be anywhere between 2-5 percent of the training set (depending on the complexity of data), we call this step as \textit{seed training}. At each round $r \in R$, we add $k$ new samples by calling the active learning functions (line \ref{alg:add_new_data}). The function first removes the chosen sample $\mathcal{S}$ from the main dataset $\mathcal{D}$ and creates $T$ instances of networks by sampling weights that was learned during the training phase. Each instance is tested on unseen data points to obtain ensemble of outputs (lines \ref{eq:ensemble_st}-\ref{eq:ensemble_end}). Depending on the type of AF (i.e., variation-ratio or entropy), the uncertainty over ensembles is calculated and the new data-points are chosen based on the largest uncertainty score. Once the model is trained on newly appended data points, the algorithm proceeds by validating on the held-out training dataset $\mathcal{D}_v$ (line \ref{alg:validation}).  Note that it is not necessary to use the validation data $D_{v}$ and it is entirely optional. We assumed that even in active learning setting, we can afford using a very small percentage of unseen data for validation. In the subsequent rounds, the best weights from the previous round are loaded (i.e.,  based on the validation set) and the algorithm resumes performing AL and re-training. Unlike \cite{beluch2018power}, after each round we \textit{do not retrain the model from scratch}. In our experiments, we found that retraining from scratch does not perform as well as re-using weights from previous rounds and re-training.

{\bf Accelerating uncertainty estimation:} As discussed earlier while the scaled normal prior BNN allows inference in a single feedforward pass but for the uncertainty estimation passing it still requires a given unlabelled example passed through multiple realization weights using the trained posterior. This is computationally expensive and we alleviate this by approximating the probability distribution of the input to the last layer directly. Equipped with this distribution we propose to directly generate the random realizations of the input before last layer non-linearity and use those to estimate the uncertainty. 

For the scaled normal posterior in \eqref{eq:BCN_Posterior} the analytical expression for the distribution is challenging due to the non-linear transformations in the neural network and dependencies due to structure of layers such as convolution layers. To illustrate this, consider a simple linear transformation of $\mathbf{x}$ by matrix $\mathbf{W} \in \mathbb{R}^{m \times n}$, bias vector $\mathbf{b}  \in \mathbb{R}^m$ given by $\mathbf{y} = \mathbf{Wx} + \mathbf{b}$ where $\mathbf{W}$ follows the posterior in  \eqref{eq:BCN_Posterior}. Each entry of $\mathbf{y}$ is a sum of $n$ independent random variables with scaled normal distribution as $y_i = \sum_{j=1}^n W_{ij}x_j + b_i$. Considering the fact that a typical neural network has multiple layers and non-linearities, the calculation of analytical distribution is non-trivial. However, under the assumption that $y_i$ is Gaussian,  the mean and variance of $y_i$ after passing through ReLU non-linearity is analytically tractable. Based on this, for a multi-layer neural network comprising of $L$ dense layers we can obtain the distribution of components vectors before the non-linearity from the distribution of layer input. Suppose weights of $l^{th}$ layer are represented by $\mathbf{W}^l \in \mathbb{R}^{n^l \times n^{l+1}}$ where $n^{l+1}, n^l$ are the input and output dimension of this layer. Then the expectation and variance of components of the vector $\mathbf{y}^{l} =\mathbf{W}^l\mathbf{x}^{l-1} + \mathbf{b}^l $ can be obtained terms of first and second order moments of components of $\mathbf{x}^{l-1}$ as follows
\begin{align}
	\mathbb{E} \left[ y_i^l \right] & = \sum_{j=1}^{n^l} \mathbb{E} \left[ w^l_{ij} \right] \mathbb{E} \left[ x^{l-1}_j\right]  + b_i^l,\\
	\mathbb{V}[y_i^l] & = \sum_{j=1}^{n^l} \mathbb{E} \left[ \left( w^l_{ij}  \right)^2 \right] \mathbb{E} \left[ \left( x^{l-1}_j \right)^2 \right]  \nonumber \\ 
	& \quad \quad \quad- \mathbb{E} \left[ x^{l-1}_j \right]^2\mathbb{E} \left[ w^l_{ij} \right]^2,	
\end{align}
where $ \mathbb{E} \left[ w^l_{ij} \right]^2 = (\mathbf\sigma_{z_j}^2+\mu_{z_j} ^2) ( \sigma_{ij}^2+ \mu_{ij}^2)$.
Further, the input to next layer is obtained passing $\mathbf{y}^{l}$ through a ReLU non-linearity as $\mathbf{x}^l = \textrm{ReLU}\left(\mathbf{y}^{l}\right)$. We assume that non-linearity until the last layer is ReLU. Finally, assuming that components of $\mathbf{y}^{l}$ are Gaussian with mean and variance computed as above, the mean and variance of components of  $\mathbf{x}^l$ are given below 
%\begin{align}
%	\mathbb{E} \left[ x_i^l \right] &=  \mathbb{E} \left[ y_i^l \right] \Phi \left( \frac{\mathbb{E} \left[ y_i^l \right]} { \mathbb{V} \left[ y_i^l \right] } \right) +\mathbb{V} \left[ y_i^l \right] f\left(- \frac{\mathbb{E} \left[ y_i^l \right]} {\mathbb{V} \left[ y_i^l \right] }  \right), \\
%	\mathbb{E}[(x_i^l)^2] &= \left (  \mathbb{E} \left[ y_i^l \right]^2 +\mathbb{V} \left[ y_i^l \right] \right) \Phi\left( \frac{\mathbb{E} \left[ y_i^l \right]} { \mathbb{V}\left[ y_i^l \right] } \right)  \nonumber \\
%	& + \sqrt{\mathbb{E} \left[ y_i^l \right]^2\mathbb{V}\left[ y_i^l \right]} f\left(\frac{\mathbb{E} \left[ y_i^l \right]} { \mathbb{V}\left[ y_i^l \right] } \right)
%%	\textrm{Var} \left[ x_i^l \right] &= \mathbb{E}[(x_i^l)^2] -  \mathbb{E} \left[ x_i^l \right]^2, 
%\end{align}
\begin{align}
	\mathbb{E} \left[ x_i^l \right] &=  \mathbb{E} \left[ y_i^l \right] \Phi ( \delta_{i}^l ) + \mathbb{V} [ y_i^l ] f (- \delta_{i}^l), \\
	\mathbb{E}[(x_i^l)^2] &=  \left(  \mathbb{E} \left[ y_i^l \right]^2 + \mathbb{V} \left[ y_i^l \right] \right) \Phi( \delta_{i}^l)   \nonumber \\   
	& \quad + \sqrt{\mathbb{E} \left[ y_i^l \right]^2\mathbb{V}\left[ y_i^l \right]} f(\delta_{i}^l)
%	\textrm{Var} \left[ x_i^l \right] &= \mathbb{E}[(x_i^l)^2] -  \mathbb{E} \left[ x_i^l \right]^2, 
\end{align}
where $\delta_i^l = \mathbb{E} \left[ y_i^l \right]/\mathbb{V} \left[ y_i^l \right] $ , $\Phi$ and $f$ are c.d.f. and p.d.f. of standard Gaussian distribution. Using above equations (6) to (9) the mean and variance at the input before non-linearity of each layer can be computed iteratively till the last layer $\mathbf{y}^L$. Equipped with this information the uncertainty can then be computed by directly generating  $\mathbf{y}^L$ and then passing them through the last layer's non-linearity.

\begin{algorithm}
\scriptsize
%\SetAlgoLined
\DontPrintSemicolon
\caption{Uncertainty-based smart data sampling }\label{alg:gen_prcss}
\SetKwProg{Acq}{ActiveSubSelectData}{}{}
\SetKwProg{Tr}{TrainModel}{}{}
{\bf Inputs}: Training dataset $\mathcal{D}$, validation dataset $\mathcal{D}_v$, test dataset $\mathcal{D}_t$,  seed dataset $\mathcal{S}$, model $\mathcal{M}$, number of epochs $E$, mini batch size $b$, number of rounds $R$, acquisition size $k$, number of instances $T$.\;
\Acq{(typ)}{
	\text{$\mathcal{D} \leftarrow \mathcal{D} - \mathcal{S}$}\;
	$ENO \leftarrow [\,]$(array to hold ensemble outputs)\; \label{eq:ensemble_st}
	\For{$j \in T$}{
		$\mathcal{M}_s \sim \mathcal{M}(\mathbf{w})$(sample from a weight instance)\;
		$ENO \leftarrow$ append the model output $\mathcal{M}_{s}(\mathcal{D})$ \label{eq:ensemble_end}
	} 
	$new \leftarrow$ get uncertainty of \textit{ENO} with eq(\ref{eq:entropy}) {\bf if} $typ$ is "Entropy" {\bf else} use variation-ratio $v$\;
	$new \leftarrow$ sort $new$ and get top-$k$ datapoints with the most uncertainty\;
	$\mathcal{S} \leftarrow$ $\mathcal{S} \cup new$\;
}

\text{Train($\mathcal{M}$)} with seed sample\; \label{alg:seed_sample}
\For {each $r \in R$}{
	{\bf ActiveSubSelectData}("Entropy/Variation-Ratio")\; \label{alg:add_new_data}
	\For {each $e \in E$}{
		
		Train($\mathcal{M}$) on $\mathcal{S}$ \;
		$v\_loss \leftarrow $ Evaluate($\mathcal{M}$) on $\mathcal{D}_v$ and get validation loss \;  \label{alg:validation}
		{\bf if} EarlyStoppingCriteria($v\_loss$): {\bf break}		
		}
	$\mathcal{M} \leftarrow$ Load best weights based on $\mathcal{D}_v$ (if continual training)
	}
Test($\mathcal{M}$) on $\mathcal{D}_t$
\label{algo:gen_process}
\end{algorithm}

\section{Experiments}

Starting from simple multi-layer perceptrons to deep CNN models we perform several experiments  to understand the effectiveness and shortcomings of BNNs over EN and MCD. Our objective is to analyze BNNs from the following perspective:
\begin{inparaenum}[(1)]
	\item overall efficiency in acquisition of new data points
	\item robustness of BNNs during minimal retraining (i.e., retraining by reusing the trained model from previous round) and
	\item impact of model capacity, and ensemble size.
\end{inparaenum} In addition to this, we also show the outcome of the proposed accelerated uncertainty estimation over dense neural networks.

\noindent {\bf Dataset and models}: The experiments are performed on four image classification datasets and one regression dataset. For classification, we chose MNIST, Fashion MNIST (FMNIST) \cite{xiao2017fashion}, CIFAR10 and CIFAR100 \cite{krizhevsky2009learning}. For regression, we chose the housing price prediction dataset introduced by \cite{ahmed2016house}. It consists of 535 unique houses sampled from the state of California. Each house is represented by both visual and textual data with the visual features representing the front side of the house, the kitchen, the bedroom and the bathroom. The following neural network architectures are used in this paper (for detailed description please refer Appendix 6.1). 
%Appendix \ref{appendix:a}
\begin{inparaenum}[(a)]
\itemsep0em
	\item \textit{LeNetD2}: a simple densely connected network with 300 and 100 neurons in the first and second layer respectively, 
	\item \textit{LeNet5} \cite{lecun1998gradient} :  consists of 2 CNN layers followed by a classifier network with three dense layers,
	\item \textit{AlexNet Light (ANL)}: this is simplified version of the Alexnet architecture that is similar to the K-CNN used in \cite{beluch2018power}. It  has 4 CNN layers followed by two dense layers,
	\item \textit{VGG}: there are several variations of VGG. For our experiments, we use VGG19 \cite{simonyan2014very} with 16 CNN layers followed by the classifier network, and
	\item \textit{Densenet}:  we use Densenet121 \cite{huang2017densely} with a growth rate of 32.
\end{inparaenum}
The Bayesian versions of these models were implemented from scratch. At the time of writing, there are no standardized module for Pytorch\footnote{ Models were implemented using Pytorch(pytorch.org) and Numpy(numpy.org) libraries} that can seamlessly convert a conventional neural network to its Bayesian counterpart.  The complete active learning library consisting of all the models and experiments performed in this paper will be publicly hosted via Github \footnote{github.com/VRM1/ActiveLearning}. 

\noindent {\bf Active learning setting}: Our training procedure was explained in Algorithm \ref{alg:gen_prcss}. Depending on the model complexity and the dataset, the number of rounds $R$ is set anywhere between $40$ to $80$. The number of data-points added during each round (i.e, $k$) varies from $5$ to $250$. The number of neural network instances (NNI) to compute uncertainty (i.e., the variable $T$) is set as $15$ for Densenet and $25$ for all other neural networks. For BNN, the instances are created by sampling from the posterior distribution of weights and for MCD, the dropouts are activated during the active learning phase (but turned off during testing phase). Similar to \cite{beluch2018power}, for EN, the number of NNI are set as 5 and the weights are initialized using the default Pytorch setting, which follows Kaiming uniform distribution. The details of neural network architecture, dataset, epochs, acquisition size, etc., are summarized in Table \ref{tab:exp_setting}. Experiments are performed by either reusing the state of the model from previous round and retraining, termed as \textit{continual training} (CT) or completely resetting the model and \textit{retraining from scratch} (RFS). In CT, in each new round, models are trained for a significantly lower number of epoch, typically around 30-50, depending on the type of model and dataset.On the contrary, in RFS the number of epochs range anywhere between 100-200, which makes the training time much greater than CT.
%Here, the Epoch column is divided into 2 parts (separated by ``\textit{/}"). The first part indicates \# epochs for CT and the second for RFS, where the first part indicates seed sample size, the second is the \# datapoints added in each round and the third is the total acquisition size across all rounds.

\noindent {\bf \small Evaluation metrics}: For classification  task, the following metrics are used
\begin{inparaenum}[(1)]
	\item \textit{Top-1 Accuracy}: is the ratio between the number of correct predictions and the total number of predictions. We take the class that corresponds to the highest probability (i.e., from the softmax layer) as the predicted label,
	\item \textit{Pr}: precision is the fraction of true positives among the retrieved instances, and
	\item \textit{F1}: F1 is the harmonic mean of the precision and recall, where recall is defined as the fraction of the total relevant instances (i.e., both true positives and true negatives) that were retrieved.
\end{inparaenum}

Aside from the aforementioned performance metrics, it is also important to know the calibration of our models since a high accuracy does not imply good calibration and vice versa \cite{guo2017calibration}. Therefore, for classification tasks, we use \textit{expected calibration error (ECE)} \cite{naeini2015obtaining} as the metric of choice.  ECE approximates the level of calibration by partitioning predictions into $M$ equally-spaced bins and taking a weighted average of the bin's accuracy/confidence difference. More precisely,
\vspace{-1mm}
\begin{align}
	ECE = \sum_{m=1}^{M} \cfrac{|B_{m}|}{n} |acc(B_{m}) - conf(B_{m})|
\label{eq:ece}
\end{align}
where n is the total number of samples and $B_m$ indicates a group of samples whose prediction confidence falls into a certain interval. $acc(B_{m})$ is the ground truth accuracy  and $conf(B_m)$ is predicted probability (termed as confidence) of $B_m$. For regression task, we use the \textit{the coefficient of determination $R^{2}$}, which is a measure of the closeness of the predicted model relative to the actual model \cite{nagelkerke1991note}.  
$R^{2}$ is defined as $1 - SSE/SST$, where $SSE = \sum_{i=1}^{n}(\hat{y}_{i} - y_i)^{2}$ and $SST = \sum_{i=1}^{n}(\bar{y} - y_i)^{2}$, $\hat{y}_{i}$ is the predicted value of $i$ and $\bar{y}$ is the observed average. 

%$1 - \cfrac{\sum_{i=1}^{n}(\hat{y}_{i} - y_i)^{2}}{\sum_{i=1}^{n}(\bar{y} - y_i)^{2}}$

% follows:
%\begin{align}
%	R^{2} = 1 - \cfrac{\sum_{i=1}^{n}(\hat{y}_{i} - y_i)^{2}}{\sum_{i=1}^{n}(\bar{y} - y_i)^{2}}
%\end{align}
%where $\hat{y}_{i}$ is the predicted value of $i$ and $\bar{y}$ is the observed average. 

\begin{table}[]
\small
\begin{tabular}{lllll}
\hline
Dataset  & Model   & Epoch(CT/RFS)  & \#Rnd & Aq Size    \\ \hline
F/MNIST & LenetD2 & 30/100 & 40    & 1K/100/4K  \\
F/MNIST & Lenet5  & 30/100 & 40    & 100/100/4K \\
CIFAR10  & VGG16   & 50/200 & 80    & 1K/250/20K \\
CIFAR100 & VGG16   & 50/200 & 80    & 1K/250/20K \\
Housing  & LenetD2 & 50     & 40    & 50/5/200   \\ \hline
\end{tabular}
\caption{\small Settings of active learning experiments for various datasets. The acquisition size (\textit{Aq Size})is the number of new datapoints added in each round and \textit{\#Rnd} is the number of rounds. Aq Size is divided into 3 parts (separated by \textit{/}), where the first indicates seed sample size, the second is the \# datapoints added in each round and the third is the total aquisition size across all rounds.}
\label{tab:exp_setting}
\end{table}

\subsection{Results} 

\begin{figure*}
\centering
	\subfloat[(a)][MNIST-LenetD2]{\includegraphics[width=0.25\textwidth]{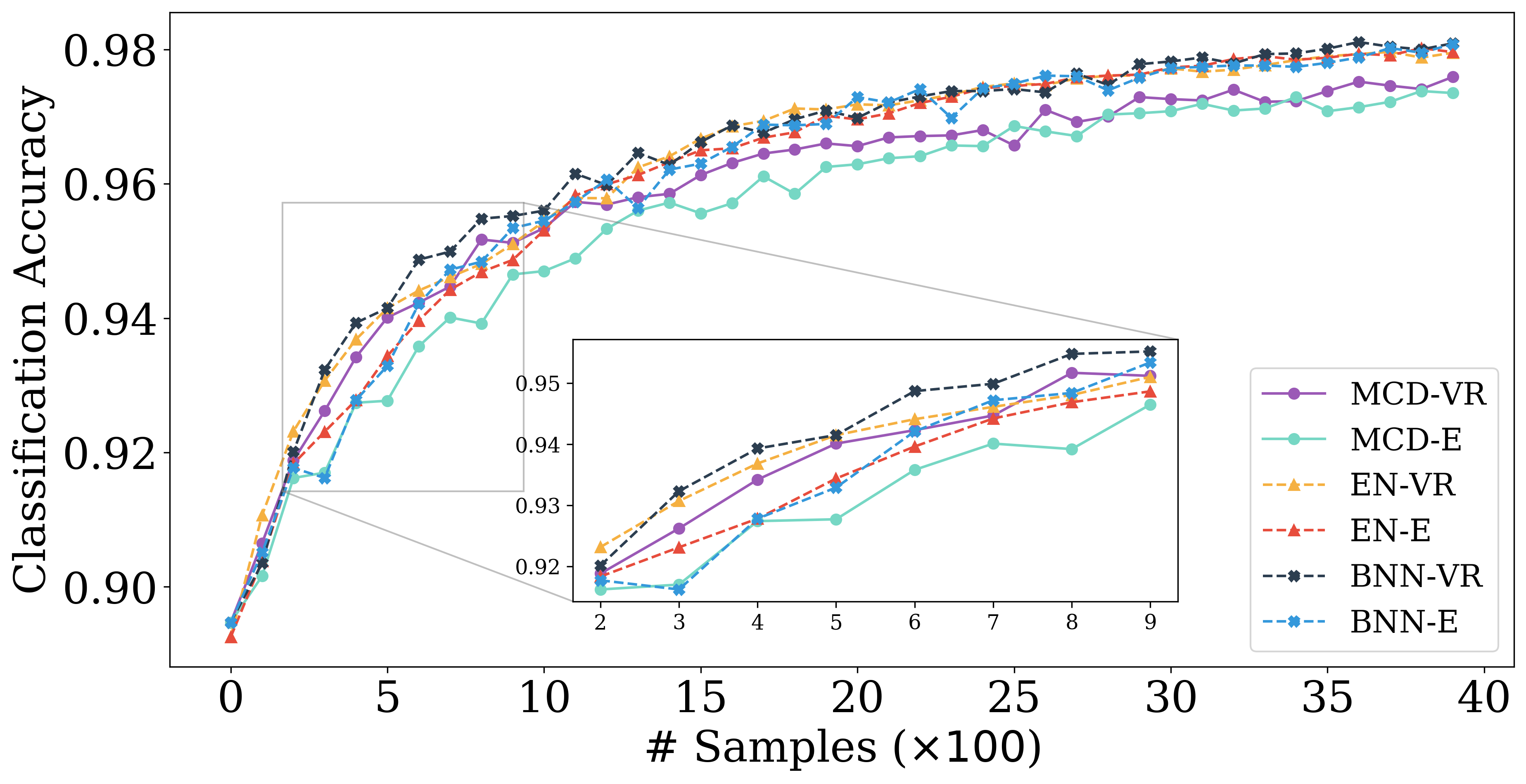}} 
	\subfloat[(b)][FMNIST-LenetD2]{\includegraphics[width=0.25\textwidth]{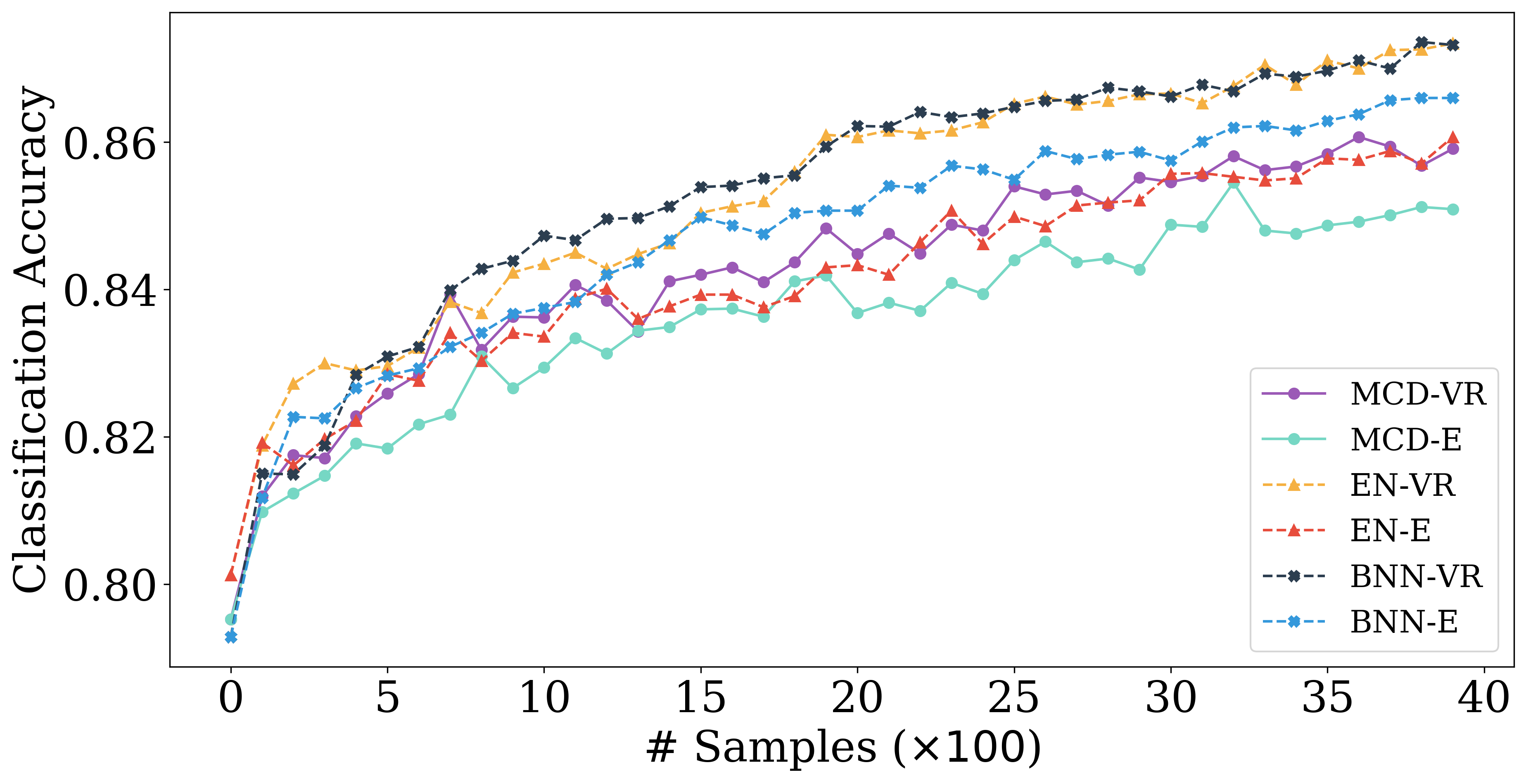}}
	\subfloat[(d)][MNIST-Lenet5]{\includegraphics[width=0.25\textwidth]{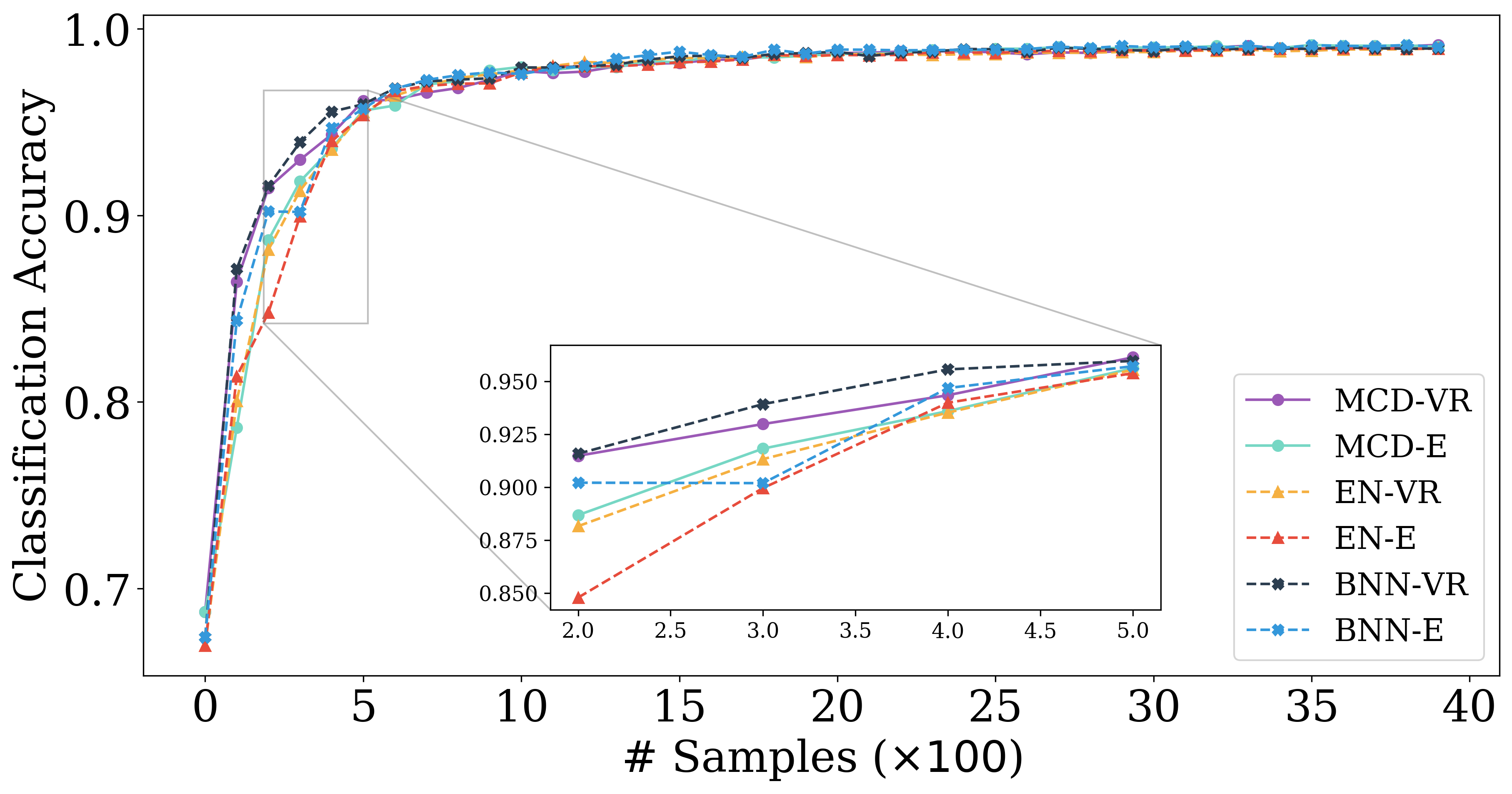}}
	\subfloat[(c)][FMNIST-Lenet5]{\includegraphics[width=0.25\textwidth]{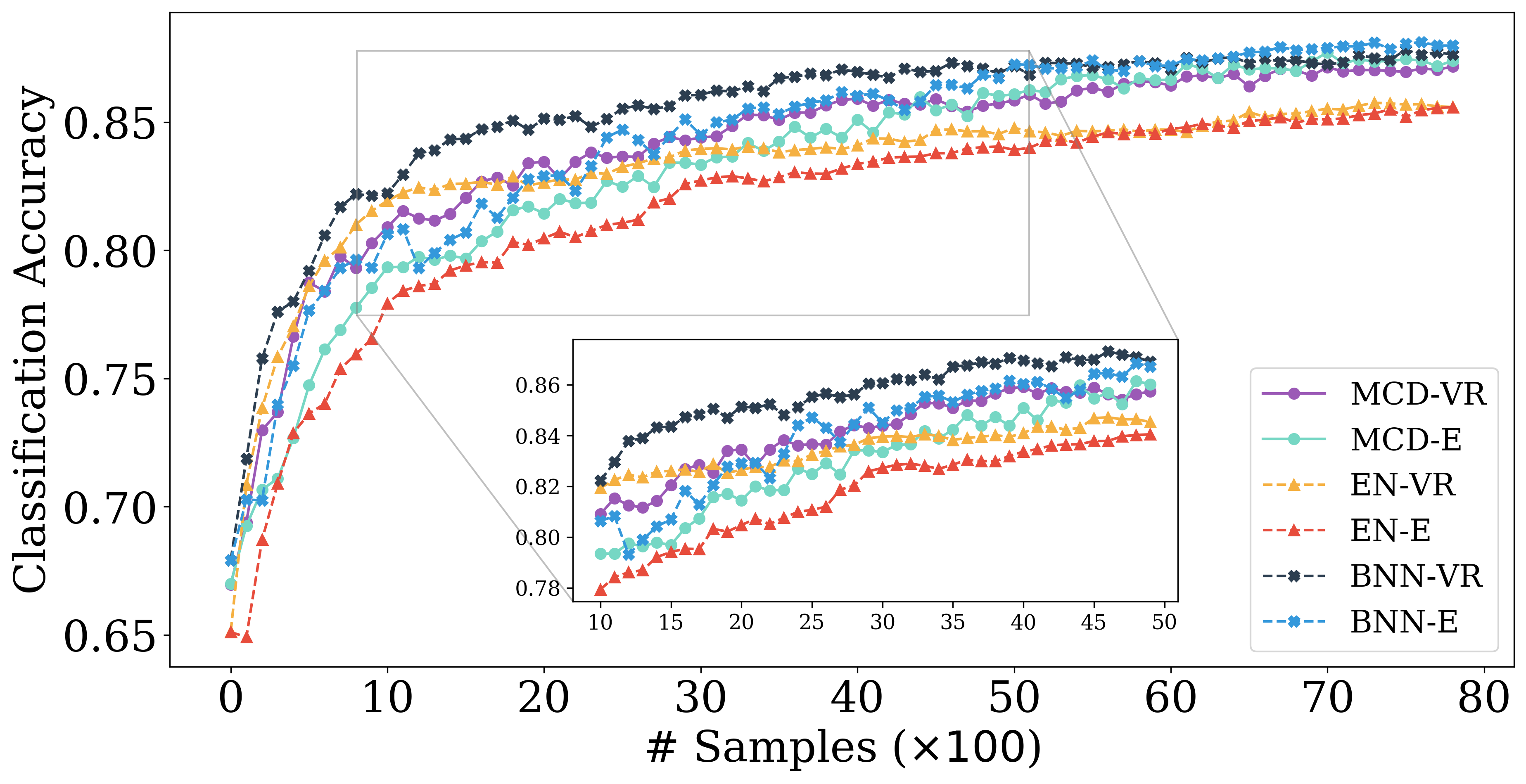}}
	\caption{Active learning performance on LeNetD2 and Lenet5 architecture with \textit{CT} setting. For MNIST, BNN and EN perform on a similar level, while MCD trails behind. For FMNIST, BNN clearly outperforms the rest and although EN performs similar to MCD upto round 10,  it clearly starts to trail behind EN and surprisingly even looses to MCD.}
	\label{fig:mnist_n_fmnist_dense}
\end{figure*}

{\bf Performance on shallow neural networks}: We begin by looking at the accuracy performance of BNN and other baselines over LeNetD2 and Lenet5 architectures. In Figure \ref{fig:mnist_n_fmnist_dense}, x-axis is the number of samples at each round, which is indicated as $(\times 100)$. For instance at $x=5$, we have added 500 samples and round 0 marks the beginning of the active learning procedure. Each model is followed by a hyphen and a letter, which indicates the type of acquisition function. For example, \textit{E} is entropy and \textit{VR} is variation ratio. For representation purpose we do not show random sampling, but in our experiments they substantially trailed behind every aquisition function.  Overall, across all models, VR tends to outperform entropy when it comes to acquisition functions. Nonetheless, this difference is more pronounced during the first half of rounds, during the final rounds the difference becomes quite narrow. This shows that EN tends to improve its acquisition quality when it sees more data. When it comes to MNIST, BNN and EN perform on a similar level, while MCD trails behind. However, due to the simplicity of dataset, the difference in performance is not that discernible. For FMNIST, which is a more challenging dataset, we start to see some interesting differences. In LenetD2  (Figure \ref{fig:mnist_n_fmnist_dense} (b)), while the performance of BNN and EN are on-par with each other (for VR), BNN seems to yield better results for entropy. In Lenet5 (Figure \ref{fig:mnist_n_fmnist_dense} (d)) although EN performs similar to BNN upto round 10, BNN clearly produces better accuracy than all other models upto round 50. Additionally, we observe EN's accuracy lagging quite significantly behind the rest after about 30 rounds. This was quite surprising to us since in \cite{beluch2018power}, the authors claim EN to perform better than MCD. Upon further investigation, we found that the poor performance of EN (compared to MCD) is \textit{only observed with CT}. In the upcoming experiments, we will show the results on both CT and RFS.

\begin{figure*}
\centering
	\subfloat[(a)][ANL-CT]{\includegraphics[width=0.25\textwidth]{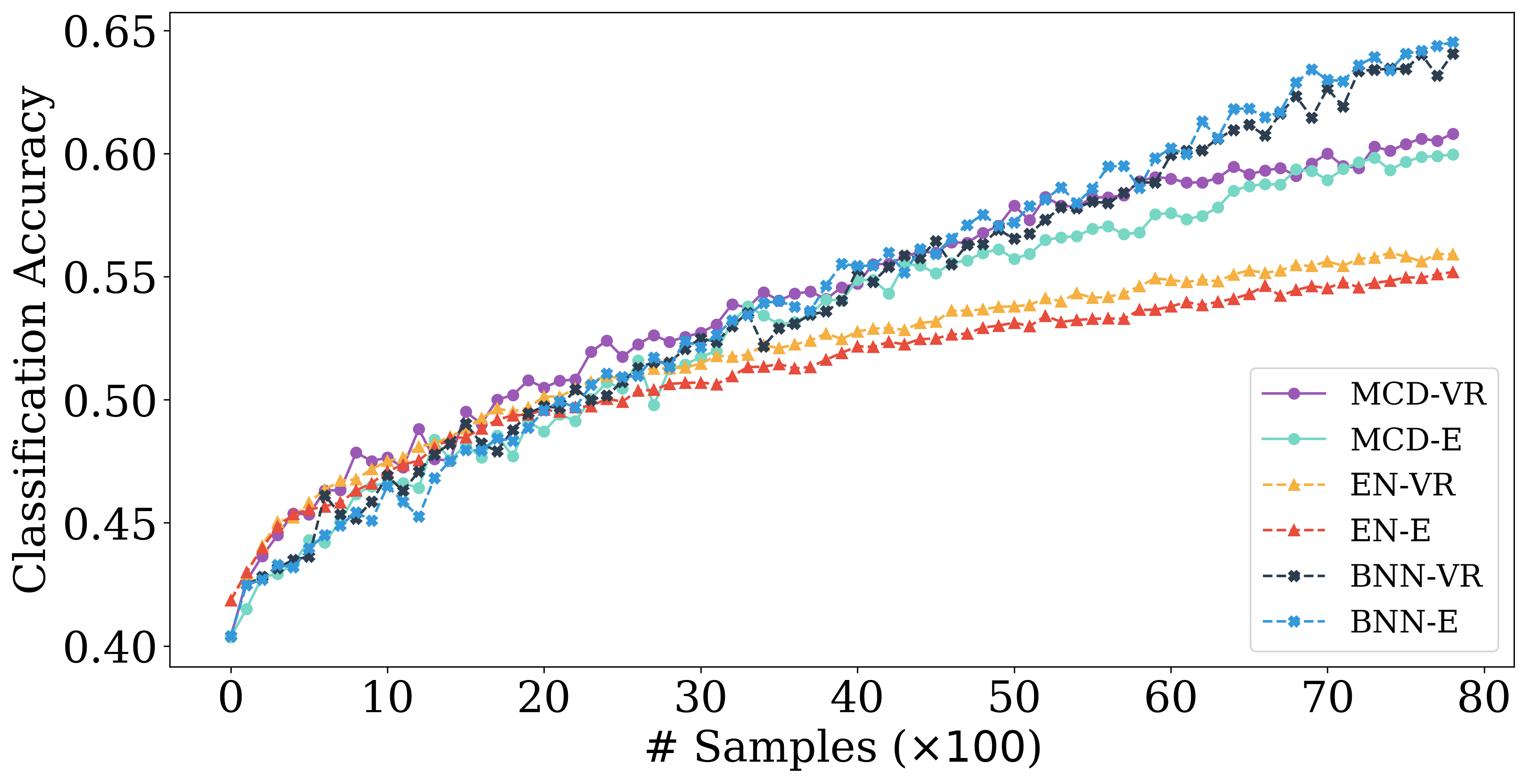}}
	\subfloat[(b)][ANL-RFS]{\includegraphics[width=0.25\textwidth]{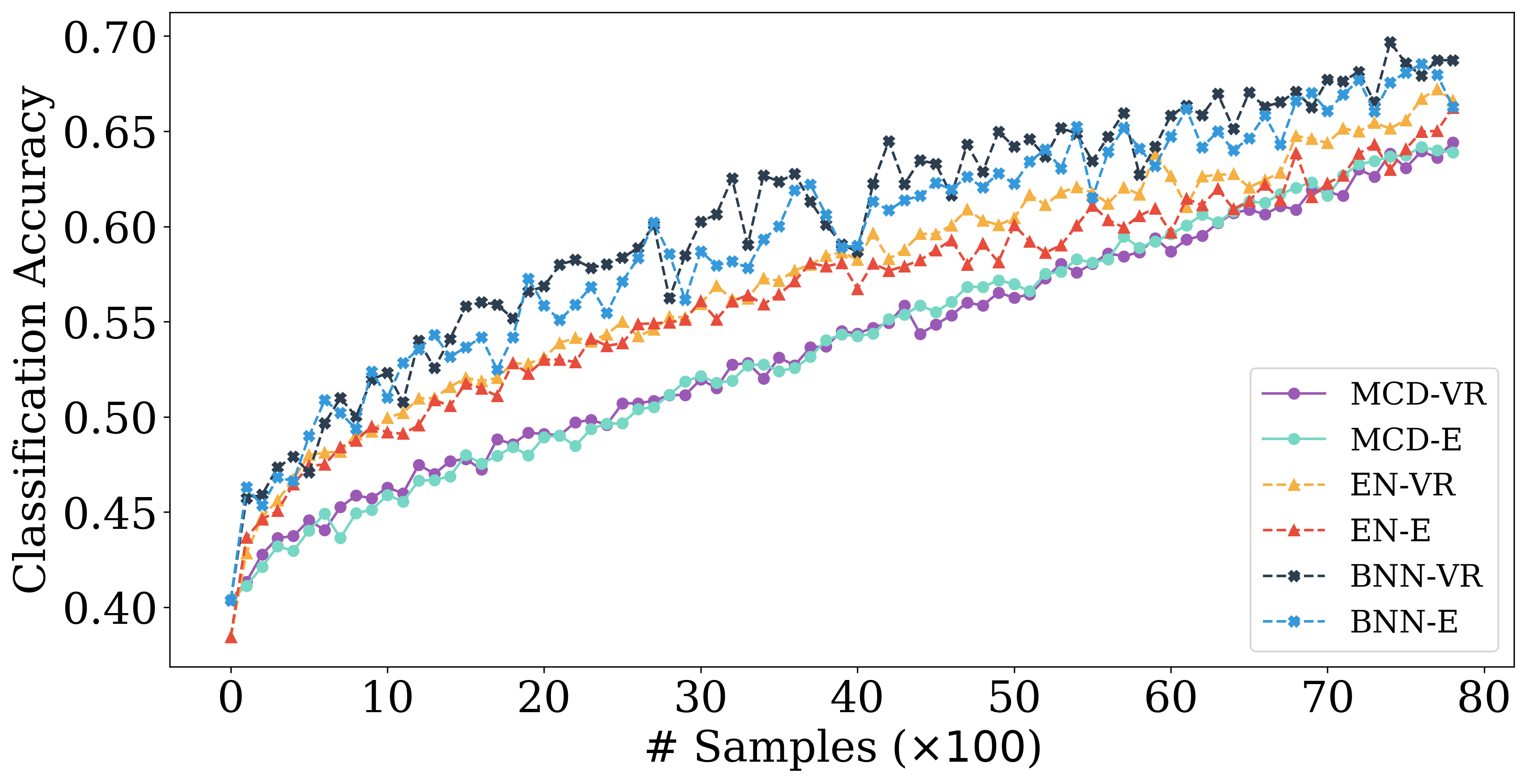}}
	\subfloat[(e)][VGG-CT]{\includegraphics[width=0.25\textwidth]{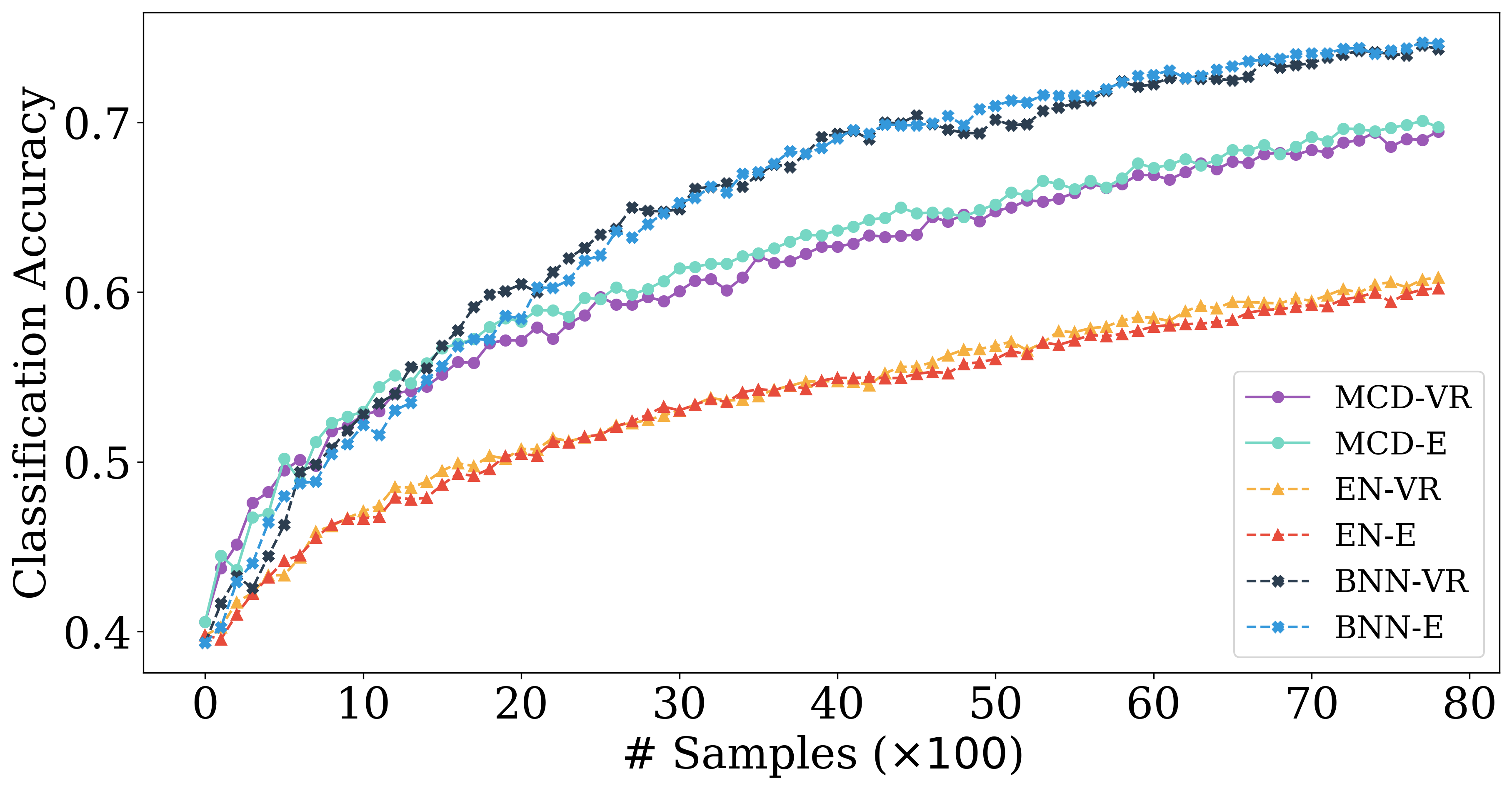}}
	\subfloat[(f)][VGG-RFS]{\includegraphics[width=0.25\textwidth]{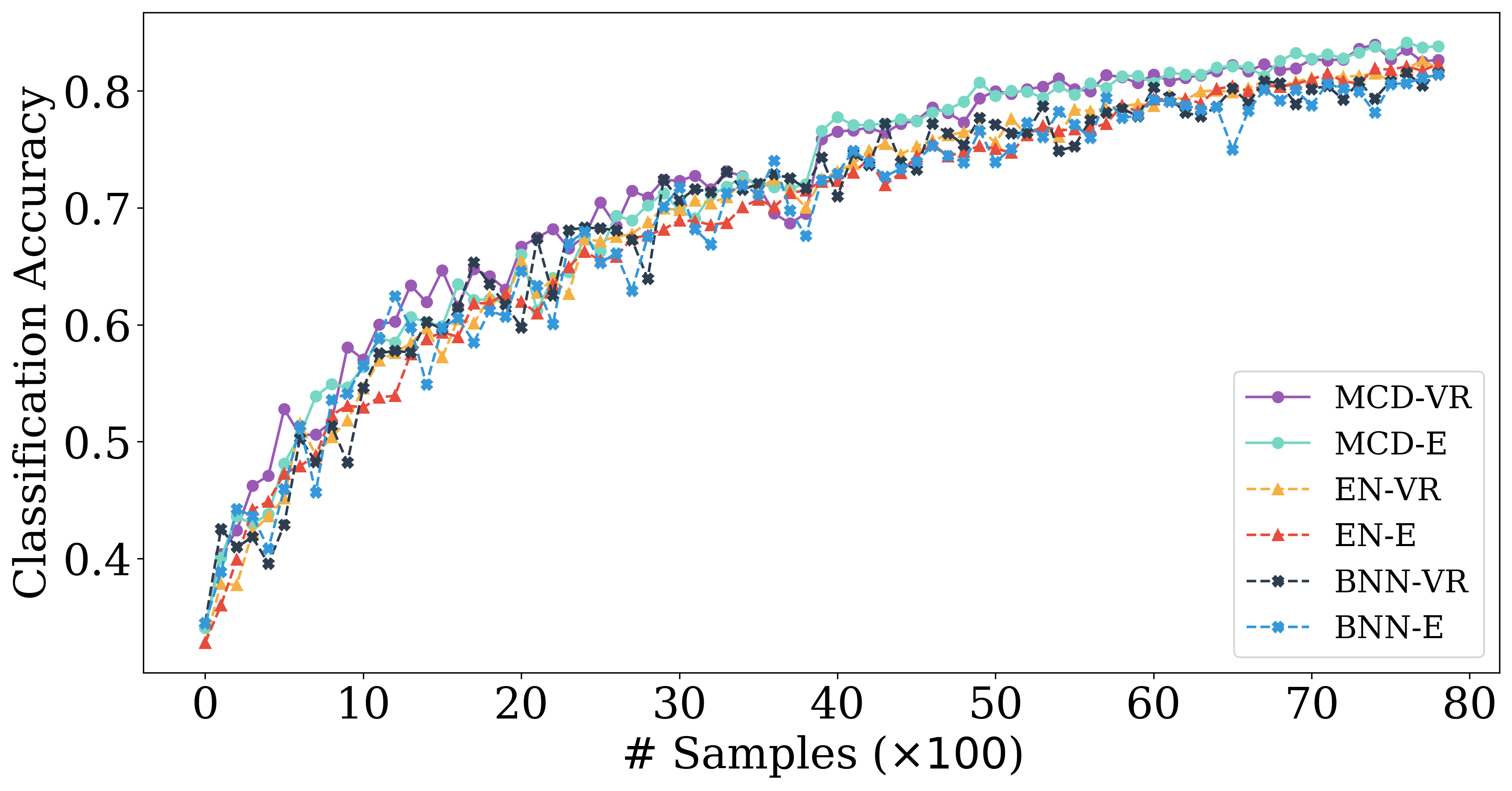}} \\
	\subfloat[(c)][ANL-CT]{\includegraphics[width=0.25\textwidth]{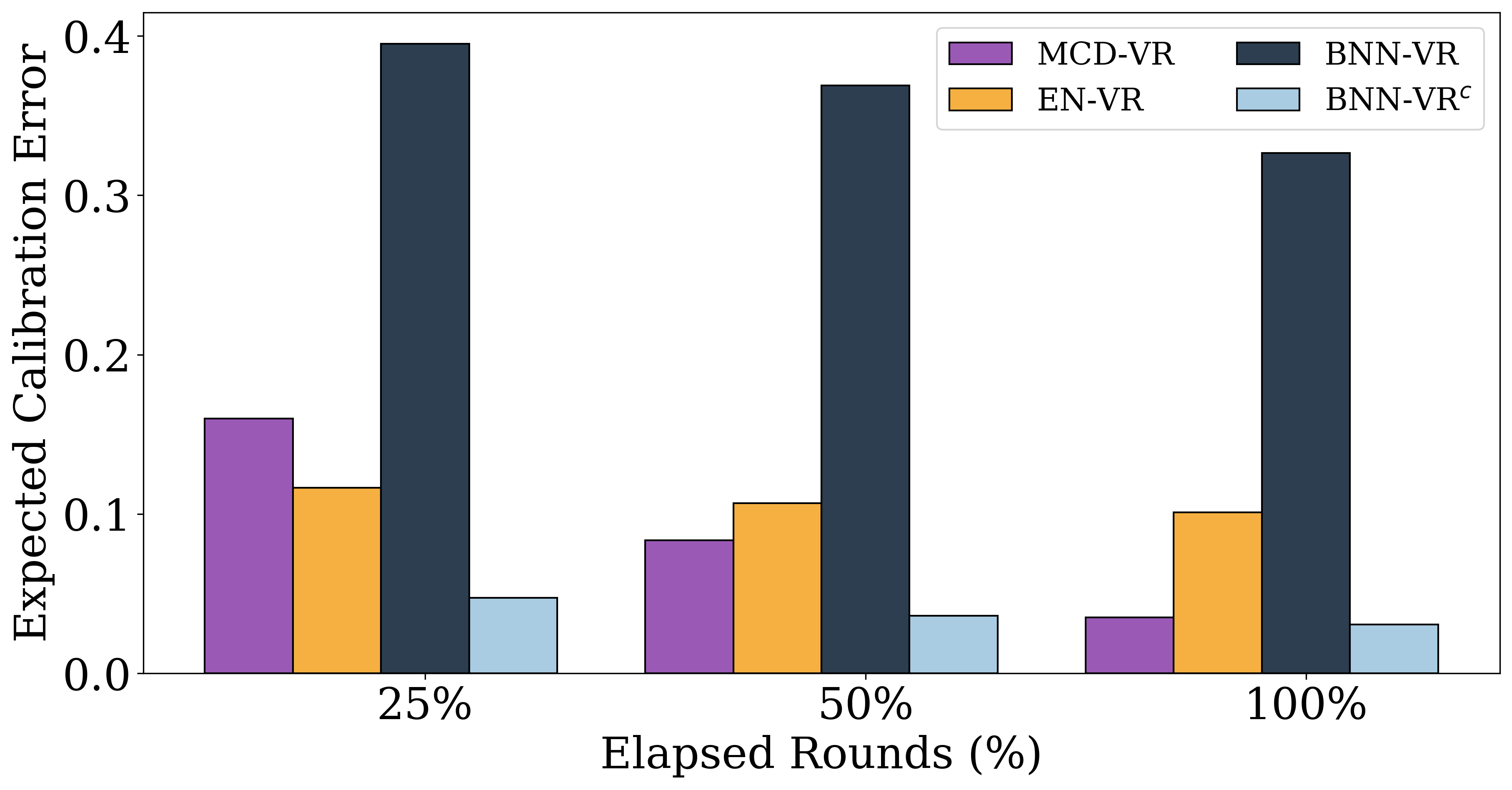}}
	\subfloat[(d)][ANL-RFS]{\includegraphics[width=0.25\textwidth]{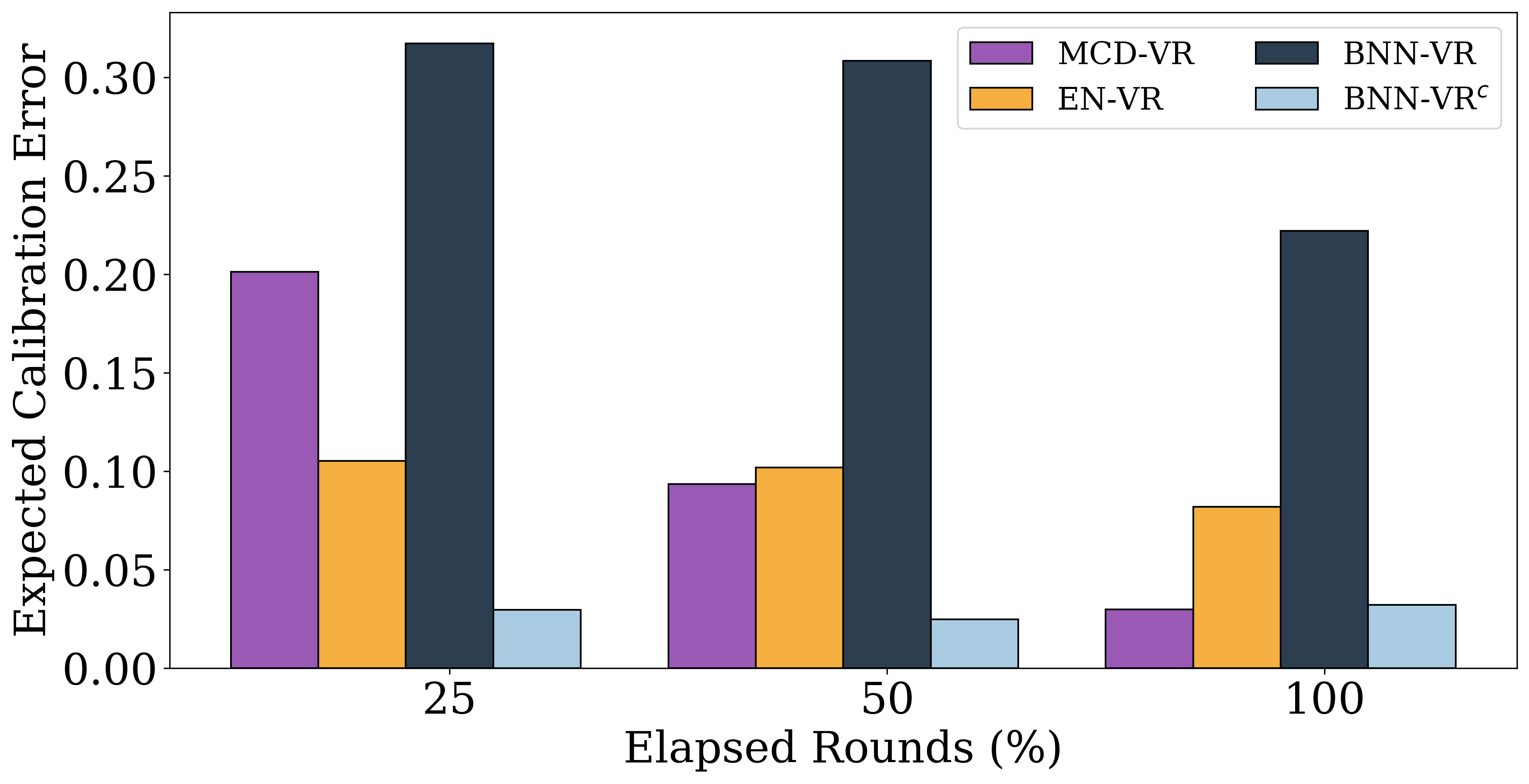}}
	\subfloat[(g)][VGG-CT]{\includegraphics[width=0.25\textwidth]{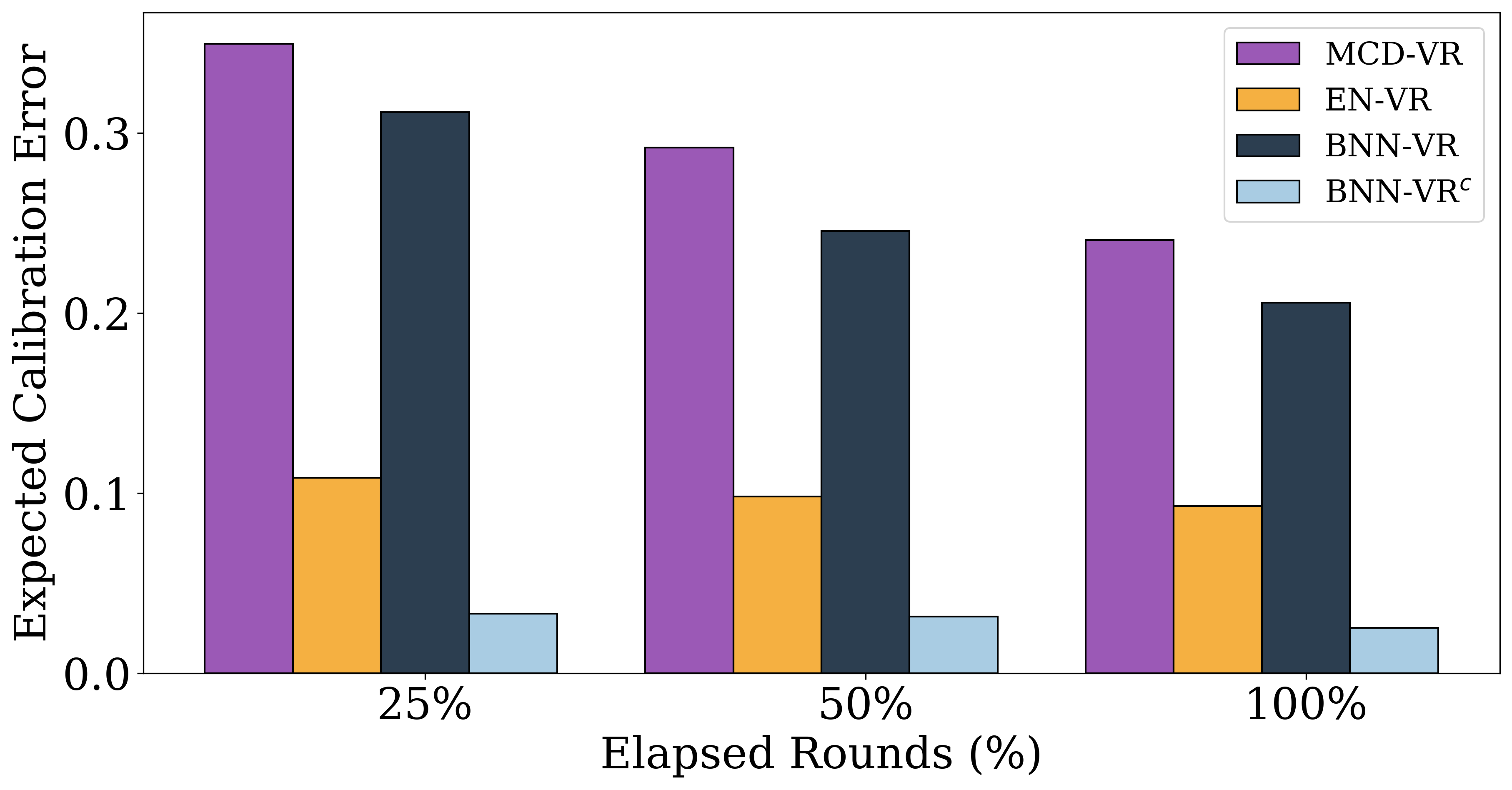}}
	\subfloat[(h)][VGG-RFS]{\includegraphics[width=0.25\textwidth]{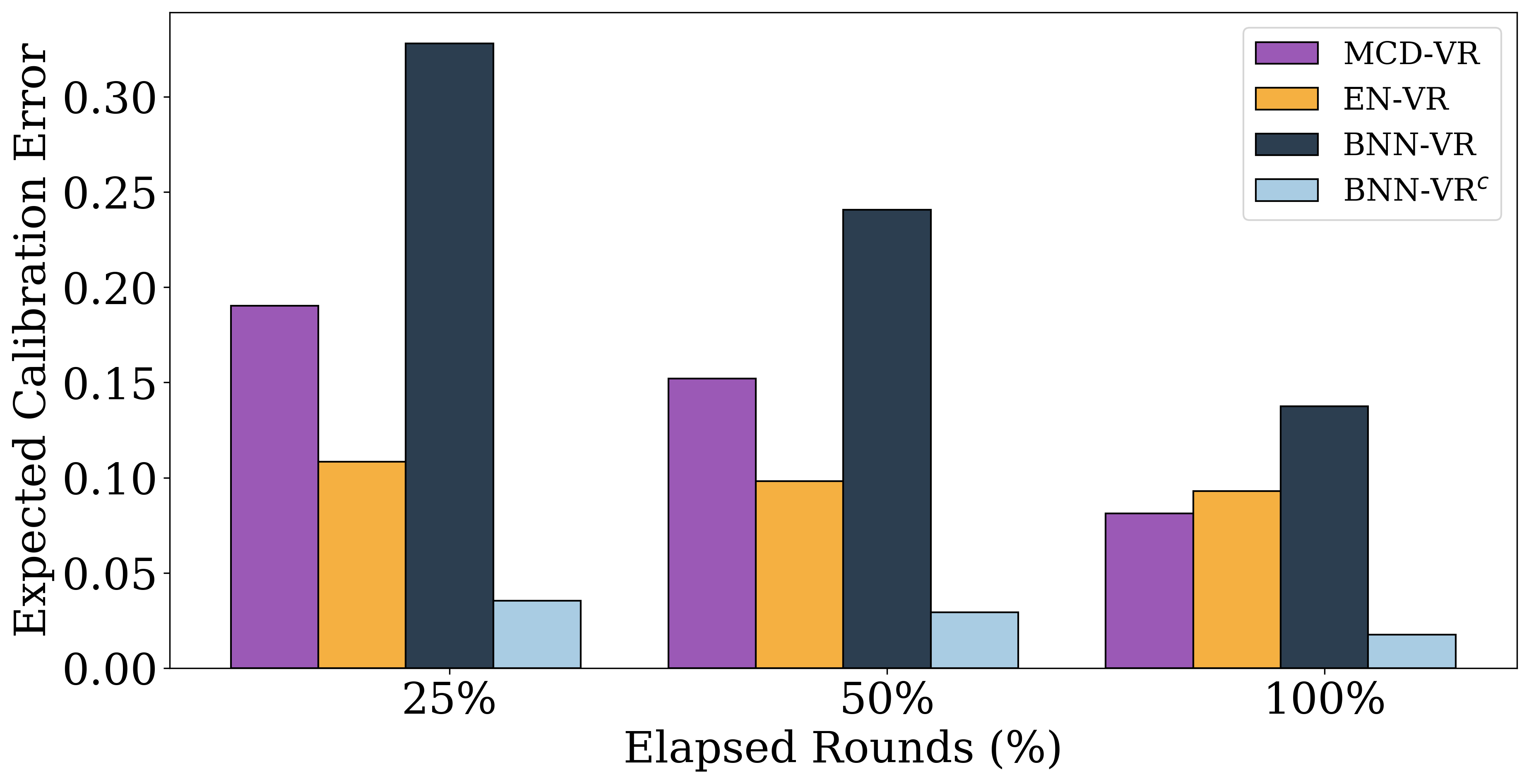}}
	\caption{Active learning performance of ANL and VGG over Cifar10 -- classification accuracy and model calibration. BNN yields a more robust performance when compared to EN and MCD, irrespective of the training methology (a)-(d), while EN suffers quite significantly in the CT setting. When retraining from scratch, EN gets a major boost  in performance and just slightly lags behind BNN for ANL and performs on par with BNN on VGG. Unfortunately, BNN suffers from poor calibration  (e)-(h), but it can be quite easily corrected. The calibrated model BNN-VR$^{c}$ is shown as the light blue bar. }
	\label{fig:anl_vgg_cifar10_accuracy} \vspace{-1em}
\end{figure*}

\noindent {\bf Performance on Cifar10}: The active learning performance of ANL and VGG is represented in Figure \ref{fig:anl_vgg_cifar10_accuracy} (a)-(d). The results unravel some important characteristics of the models. First, when it comes to CT, BNNs performance is significantly better than both EN and MCD, especially for VGG. Here, BNN achieves about 75\% and MCD about 67\% accuracy, but similar to previous results (i.e., Lenet5) EN clearly under performs with just 55\%. A possible reason for this outcome could be the lack of regularization. ENs are full capacity models without any dropouts and we observe quite substantial overfitting during CT setting. This in-turn result in making incorrect decisions when acquiring new data points during the active learning phase. Another reason could be catastrophic forgetting, which is a well known problem in continual training of neural networks \cite{goodfellow2013empirical, kemker2017measuring}. BNN on the other hand seems to be more robust to such perturbations. That being said, when retraining from scratch EN starts to outperform MCD as it doesn't overfit the data acquired in previous arounds, which aligns with the recent study \cite{beluch2018power}. For ANL (Figure \ref{fig:anl_vgg_cifar10_accuracy} (b)) we clearly see EN outperforming MCD, however, BNN still achieves better performance. When it comes to VGG (Figure \ref{fig:anl_vgg_cifar10_accuracy} (c)) we don't not find any distinct performance gaps between all three models. Overall, when retrained from scratch, all models perform better compared to CT, but we observe a lot more variations in the accuracy scores, while in CT the increase in accuracy is quite smooth. 

\noindent {\bf Calibration characteristics}: When it comes to measuring robustness of machine learning models, relying solely on performance metrics such as accuracy is not sufficient. In our experiments, the calibration score is measured as expected calibration error (ECE) using equation \eqref{eq:ece}. In Figure \ref{fig:anl_vgg_cifar10_accuracy} (e)-(h), the y-axis is the ECE and the x-axis is the elapsed round number. For instance, for both ANL and VGG since the active learning is performed until round 80 (Figure \ref{fig:anl_vgg_cifar10_accuracy} (a)-(d)), 25\% implies the 20th round of the corresponding experiment. From calibration plots, we observe that ENs have the best out-of-the-box ECE scores followed by MCD. Although BNN offers great performance in terms of accuracy, they seem to be quite poorly calibrated. Thankfully, calibration is a post-training step and there are several ways to calibrate a model such as histogram binning and temperature scaling; in this paper we adopt the later. The ECE of the calibrated BNN model is shown as the blue bar. It is important to note that even though we have significantly reduced the calibration error, the accuracy of BNN still remains unaltered. 

\begin{figure}
\centering
	\subfloat[(a)][Densenet-CT]{\includegraphics[width=0.23\textwidth]{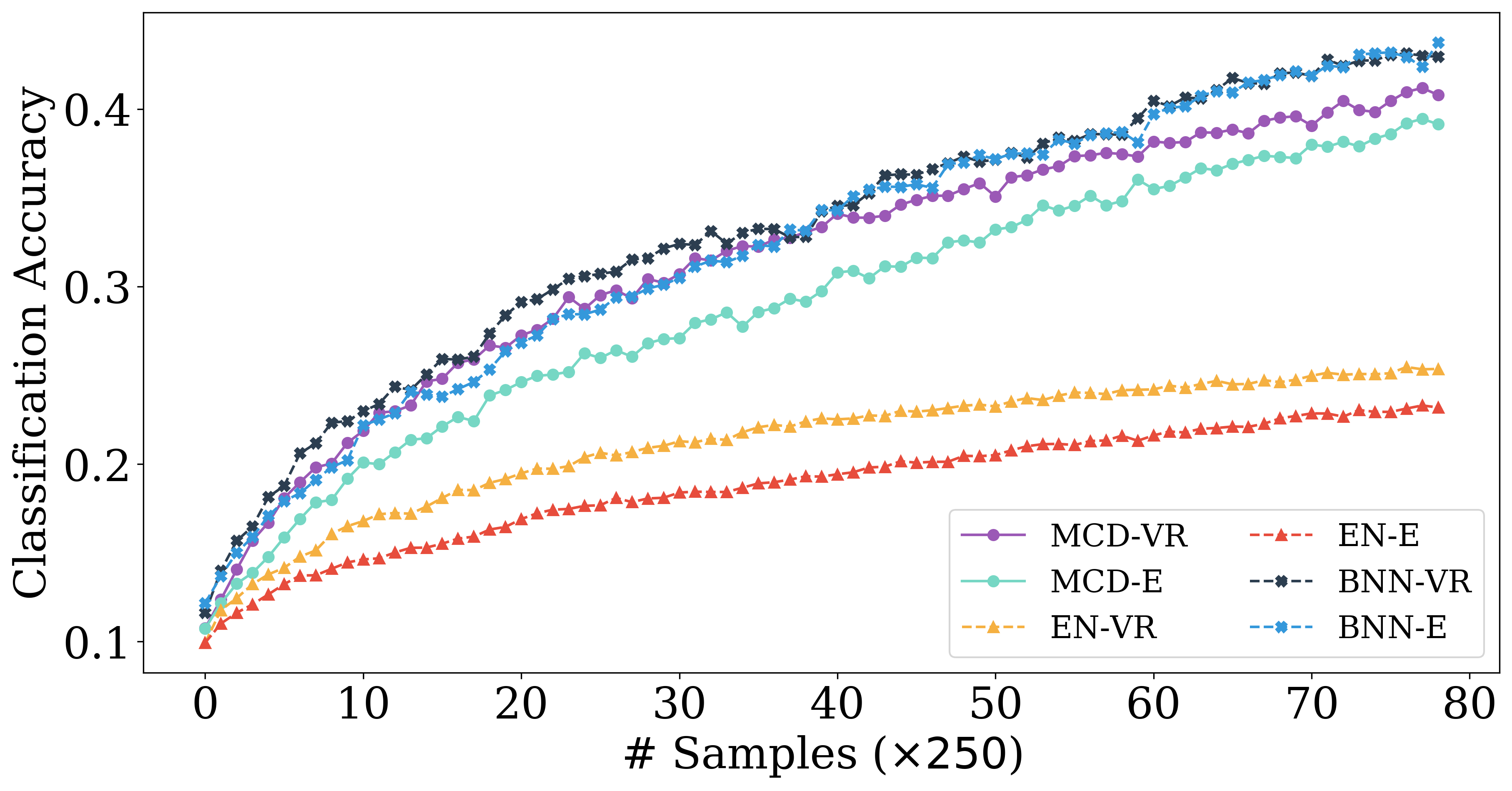}}
	\subfloat[(b)][Densenet-retrain from scratch]{\includegraphics[width=0.23\textwidth]{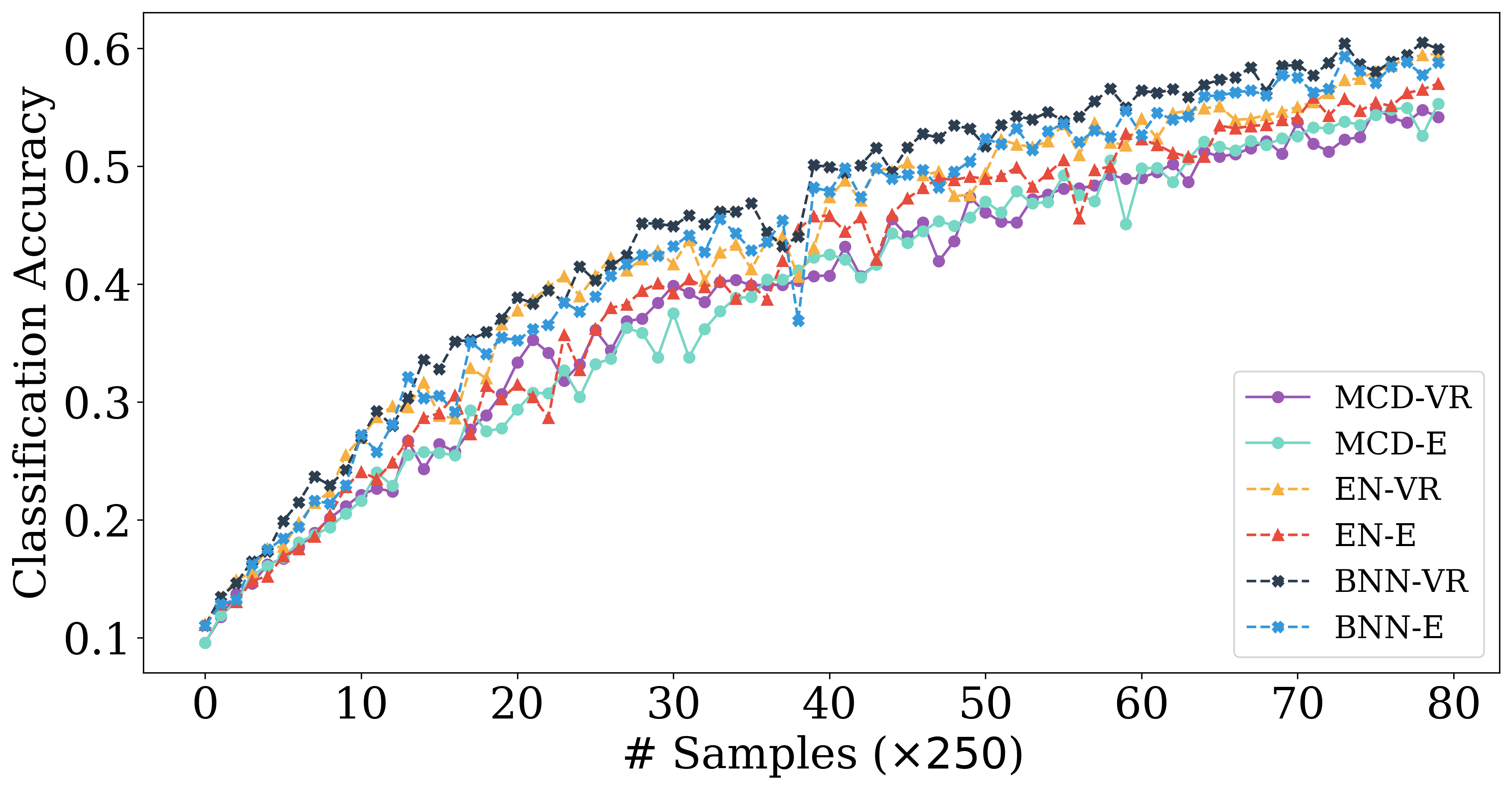}}
	\caption{Active learning performance of Densenets on Cifar100 dataset. Similar to the results on Cifar10, BNN produces a more robust performance than EN and MCD whether it is CT or retraining from scratch.}
	\label{fig:cifar100_densenet}\vspace{-1em}
\end{figure}

% a more challenging dataset compared to FMNIST.

\noindent {\bf Performance on Cifar100}: From Figure \ref{fig:cifar100_densenet}, one can observe that even for larger neural networks such as Densenets, BNN produces significantly better accuracy compared to MCD and EN. As for EN, it looks like as the neural networks become more complex, the CT methodology starts to substantially cripple the performance. 
When RFS both BNN and EN have similar performance, while MCD trails behind. 

\begin{figure}
\centering
	\includegraphics[width=0.3\textwidth]{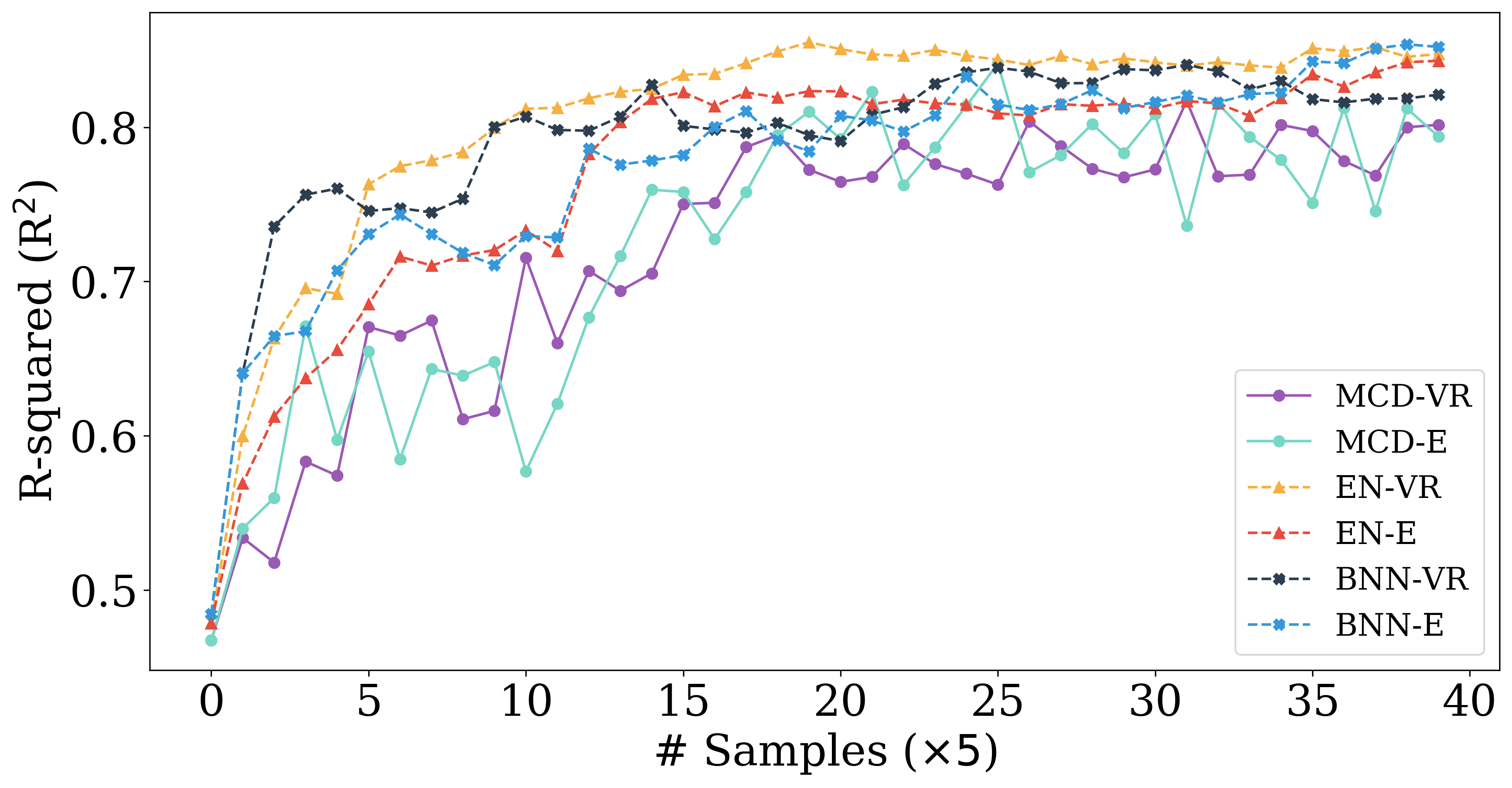}
	\caption{Housing price prediction dataset using LenetD2. There is no major performance difference between BNN and EN, but MCD clearly under performs.}
	\label{fig:regression}\vspace{-1em}
\end{figure}

\noindent {\bf Performance on regression dataset}: The robust performance of BNN is not just restricted to classification, but also extends to regression. Figure \ref{fig:regression} shows the performance of BNN in terms of $R^{2}$ on housing price prediction. For this dataset, we used the LenetD2 architecture where the input layer is a feature vector that is from by concatenating the image and descriptive text features of the individual houses \cite{ahmed2016house}. For this dataset, we start with a seed sample of 50 and in each round we add five new data points that is selected via active learning. Here there is no major performance difference between BNN and EN, but MCD clearly under performs. Additionally, when it comes to BNN there seems to be a lot more variance in $R^{2}$ score in each round, while EN tends to be smoother.
% We tried both retraining from scratch and CT and found no major difference. 

\subsection{Ablation Study}

\begin{table*}
\centering
\small
\resizebox{\textwidth}{!}{%
\begin{tabular}{l|c|c|l|c|c|l|l|c|l|c|l|c|l|}
\cline{2-14}
\multirow{3}{*}{} & Class & \multicolumn{6}{c|}{BNN-VR} & \multicolumn{6}{c|}{EN-VR} \\ \cline{2-14} 
 & \multirow{2}{*}{} & \multicolumn{2}{c|}{25\%} & \multicolumn{2}{c|}{50\%} & \multicolumn{2}{l|}{100\%} & \multicolumn{2}{c|}{25\%} & \multicolumn{2}{c|}{50\%} & \multicolumn{2}{c|}{100\%} \\ \cline{3-14} 
 &  & Pr & F1 & Pr & F1 & Pr & F1 & Pr & F1 & Pr & F1 & Pr & F1 \\ \hline
\multicolumn{1}{|c|}{\multirow{3}{*}{\begin{tabular}[c]{@{}c@{}}Cifar10\\ (VGG)\end{tabular}}} & Bird & \textbf{51.2} & \textbf{47.9} & \textbf{63.8} & \textbf{67.1} & \textbf{79.3} & \textbf{76.9} & 48.8 & 46.2 & 62.8 & 63.4 & 75.7 & 75.2 \\ \cline{2-14} 
\multicolumn{1}{|c|}{} & Cat & \textbf{50.6} & \textbf{45.4} & \textbf{59.6} & \textbf{57.8} & 67.3 & \textbf{68.4} & 49.3 & 44.9 & 58.4 & 55.6 & \textbf{69} & 68.2 \\ \cline{2-14} 
\multicolumn{1}{|c|}{} & Dog & 36.4 & 44.0 & \textbf{62.7} & \textbf{65.1} & \textbf{77.8} & \textbf{77.2} & \textbf{54.1} & \textbf{58.2} & 60.9 & 64.9 & 74 & 74.9 \\ \hline
\multicolumn{1}{|l|}{\multirow{2}{*}{\begin{tabular}[c]{@{}l@{}}FMNIST\\ (Lenet5)\end{tabular}}} & Pullover & \multicolumn{1}{l|}{\textbf{74.7}} & \textbf{75.3} & \multicolumn{1}{l|}{\textbf{81}} & \multicolumn{1}{l|}{\textbf{79.8}} & \textbf{81.8} & \textbf{80.2} & \multicolumn{1}{l|}{71.4} & 72.4 & \multicolumn{1}{l|}{74.4} & 75.2 & \multicolumn{1}{l|}{78.3} & 77 \\ \cline{2-14} 
\multicolumn{1}{|l|}{} & Shirt & \multicolumn{1}{l|}{\textbf{57.1}} & \textbf{62.8} & \multicolumn{1}{l|}{\textbf{62.7}} & \multicolumn{1}{l|}{\textbf{64.5}} & \textbf{66} & \textbf{66.3} & \multicolumn{1}{l|}{56.6} & 56.2 & \multicolumn{1}{l|}{61.9} & 60.5 & \multicolumn{1}{l|}{62.7} & 63.4 \\ \hline
\end{tabular}%
}
\caption{Precision and F1 score of BNN and EN on challenging class labels. The second row indicates the elapsed round and Pr indicates precision. Even though for VGG, EN and BNN have similar accuracy (Figure \ref{fig:anl_vgg_cifar10_accuracy}(d)), the precision and F1 measure on challenging labels indicate that performing AL via BNN leads to better results.}
\label{tab:f1_precision}
\end{table*}

We perform data-centric and model-centric ablation studies to get a deeper understanding of BNN's performance. 

\noindent {\small \bf  Performance on challenging class labels}: When retraining from scratch, we see that the performance gap between BNN and EN is quite narrow, especially on VGG (Figure \ref{fig:anl_vgg_cifar10_accuracy}(d)). Nonetheless, accuracy is a global score and when it comes to AL, it is important to measure the performance on labels that are hard to classify. Table \ref{tab:f1_precision} shows the precision and F1 measure of FMNIST and Cifar10 datasets for classes that are most challenging to predict. The corresponding accuracy plots can be seen from Figure \ref{fig:mnist_n_fmnist_dense}(d) and Figure \ref{fig:anl_vgg_cifar10_accuracy}(d) respectively. Even though BNN and EN produce similar accuracy, for more challenging class labels, the former outshines the later. Across all selected rounds of active learning, both precision and F1 measures of BNN are clearly better than EN. This shows that BNN's uncertainty estimation is more effective in acquiring datapoints that improve the model performance on challenging class labels.

\begin{figure}
\centering
	\subfloat[(a)][ANL-CT]{\includegraphics[width=0.25\textwidth]{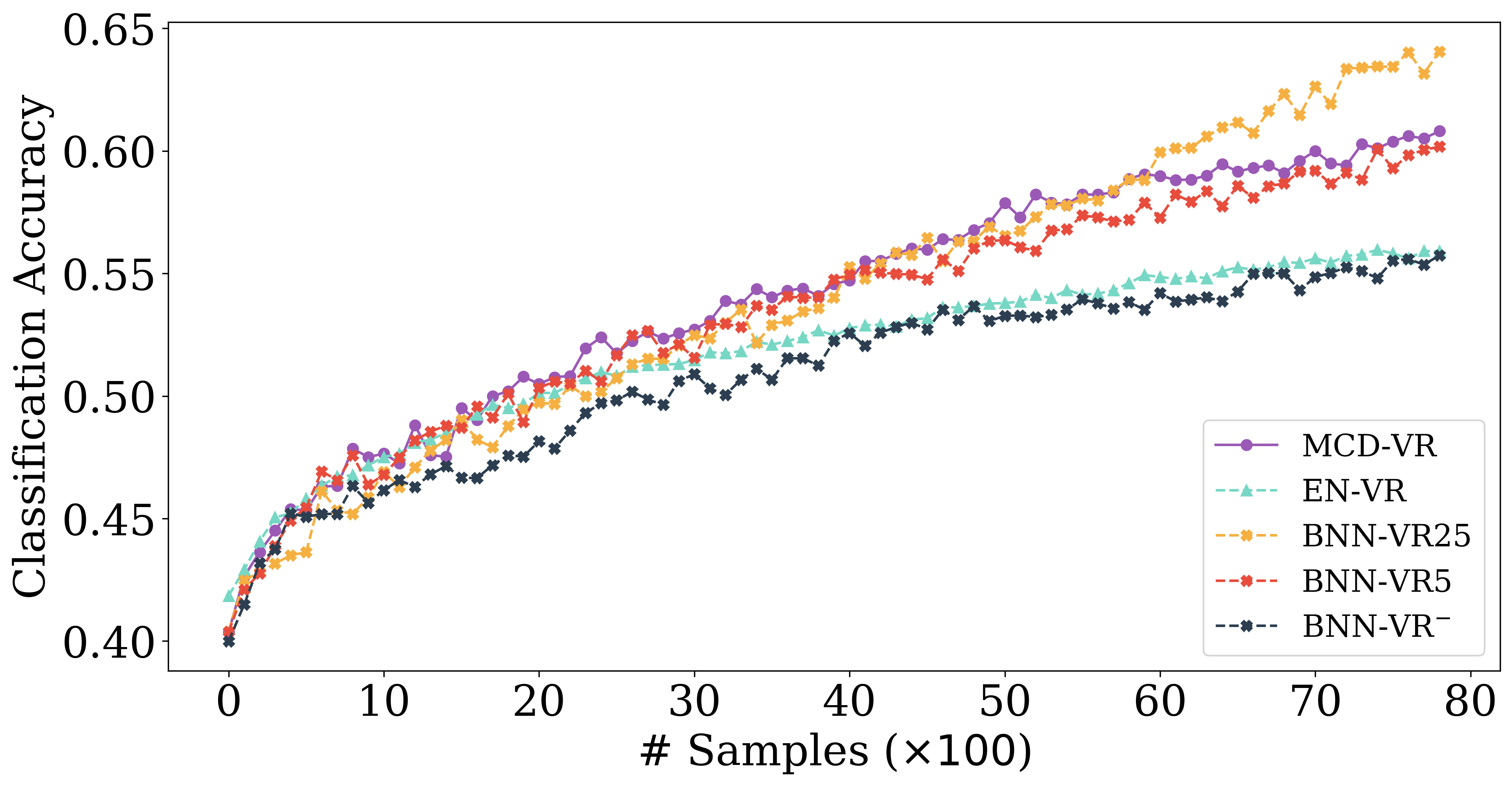}}
	\subfloat[(b)][ANL-RFS]{\includegraphics[width=0.25\textwidth]{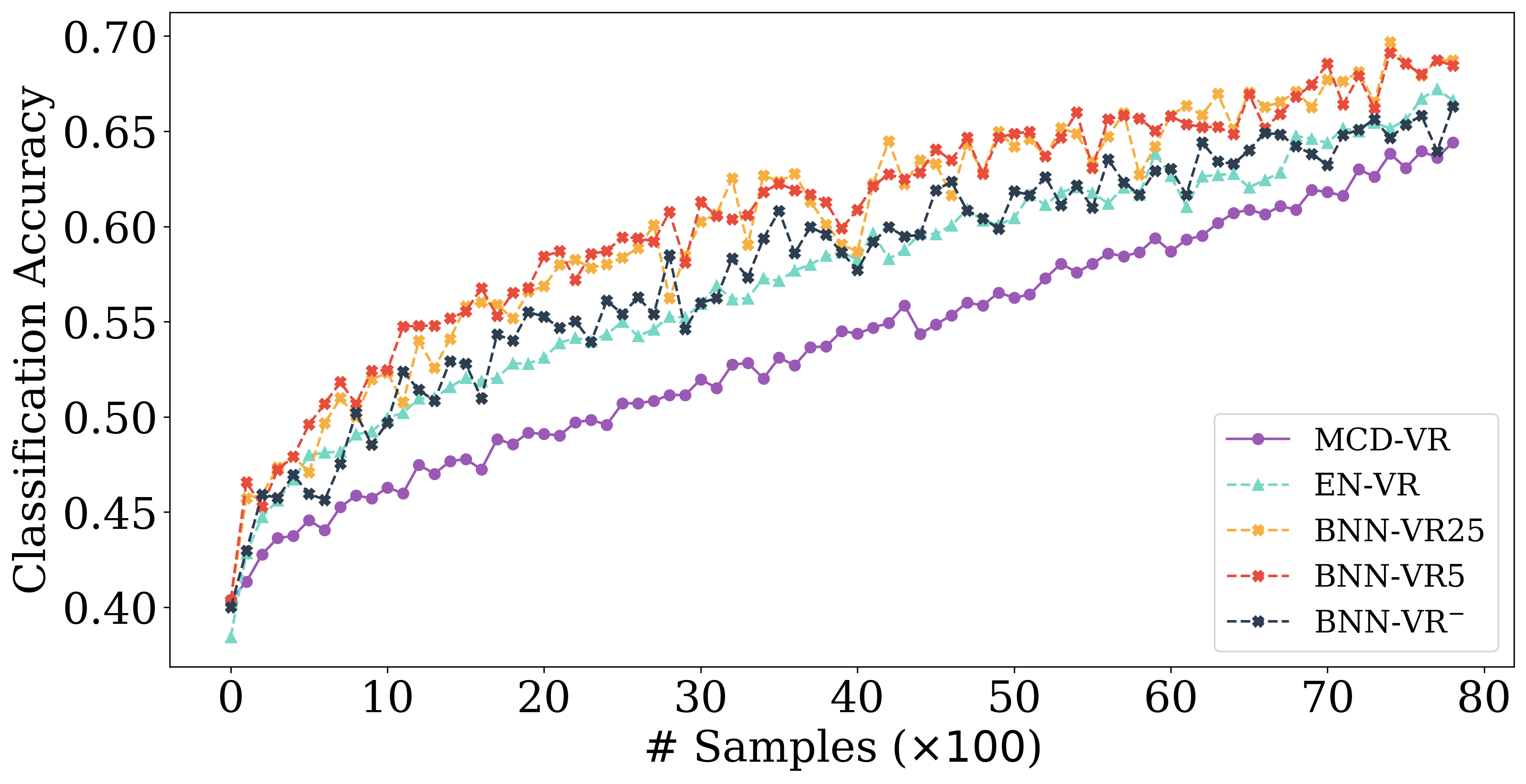}}
	\caption{Impact of number of NNI and model capacity on Cifar10. BNN-VR25 indicates 25 network instances used to estimate the uncertainty, while BNN-VR5 implies five. BNN-VR$^{-}$ is the ANL network with reduced capacity that matches those of MCD when dropouts are activated.}
	\label{fig:ablation_study_1}
\end{figure}

\noindent {\bf Impact of NNI}: The performance of Al is based on how robust is the uncertainty estimation. When it comes to BNN, a key factor that determines the goodness of estimation is the number of NNI (i.e., lines 5-7 of Algorithm 1). A major advantage of EN over BNN (and MCD) is the low number of NNI that is needed to estimate uncertainty, which is set as five in our experiments. On the other hand, for BNNs we have 25 instances. Therefore, we wanted to test how good is BNN when the NNI is reduced to that of EN. The results of this experiment is shown in Figure \ref{fig:ablation_study_1}. Here, one can observe that during CT, there is some performance dip but it is mostly towards the final rounds. On the other hand, for RFS, there is no noticeable loss in accuracy. We observed similar results even with entropy as the acquisition function. A possible reason for this outcome could be attributed to Bayesian's ability to learn distribution over weights (instead of point estimate), which naturally models uncertainty during the training process (which EN and MCD lacks). 

\noindent {\bf Impact of reduced model capacity}: Besides NNI, another factor that determine the performance of AL is the model capacity. Both BCN and EN enjoy the benefit of being a full capacity model (during training). However, for MCD, due to dropouts a significant portion of neurons in dense and CNN layers remain inactive, which is one of main reasons for MCD's inferior performance. In fact, \cite{beluch2018power} show that when ENs are capacity-limited, their performance drops roughly to that of MCD. To see this effect on BCN, we reduced the number of CNN filters and dense layers to that of MCD (i.e., 50\% less for dense and 25\% less for CNN). The outcome of this experiment is shown in Figure \ref{fig:ablation_study_1}, where BNN-VR$^{-}$ is the capacity-limited BCN. While we do see a noticeable performance dip when compared to the regular BNN, the accuracy still manages to be on-par with EN for both CT and RFS.

\begin{figure}
\centering
	\subfloat[(a)][MNIST]{\includegraphics[width=0.23\textwidth]{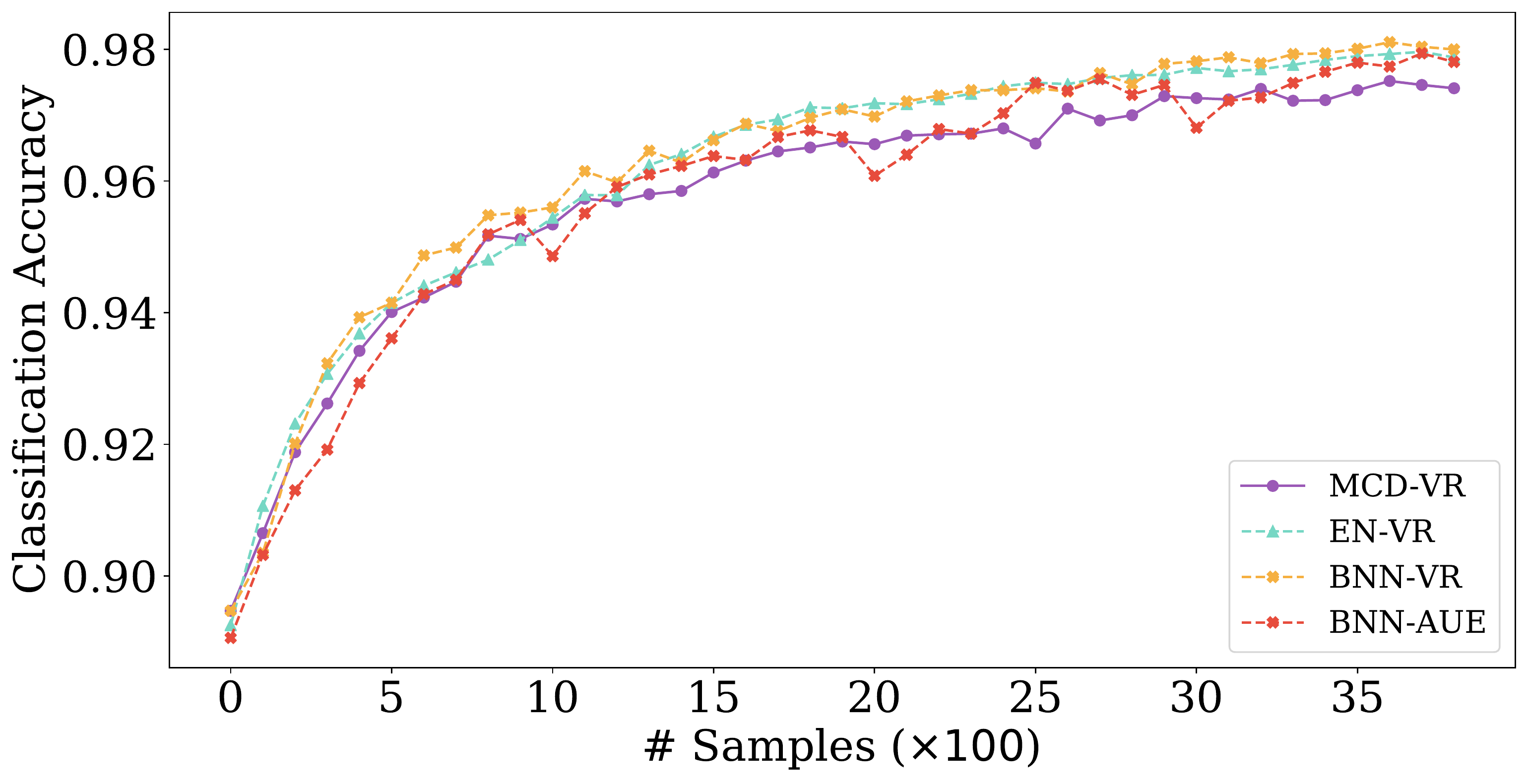}}
	\subfloat[(b)][FMNIST]{\includegraphics[width=0.23\textwidth]{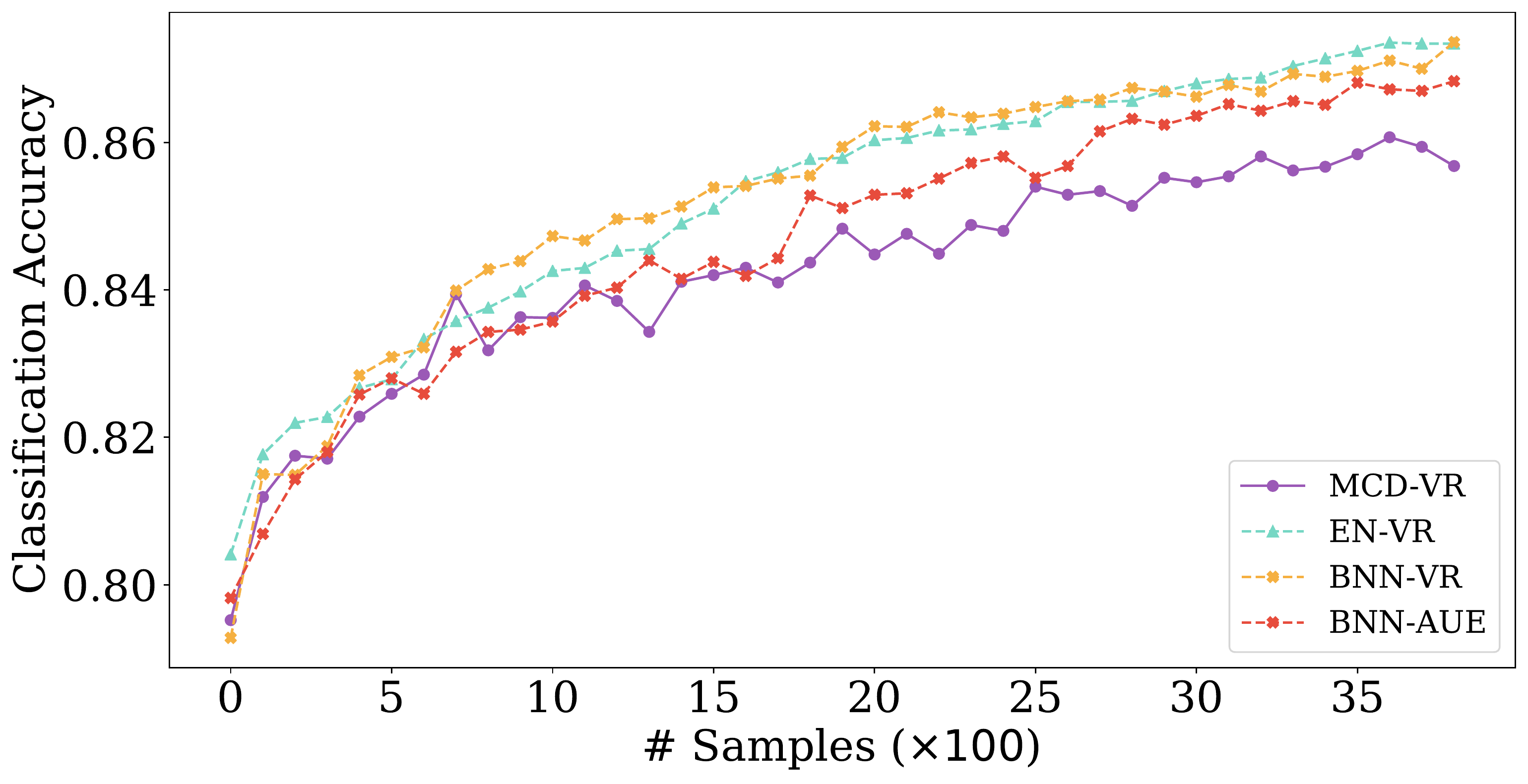}}
	\caption{Active learning performance of iterative uncertainty estimation Vs accelerated uncertainty estimation (BNN-AUE) on MNIST and FMNIST datasets.}
	\label{fig:one_shot_estimation}\vspace{-1em}
\end{figure}

\noindent {\bf Performance of accelerated uncertainty estimation}: Finally, we compare the performance of the proposed Bayesian accelerated uncertainty estimation learning (AUE) on LenetD2 architecture in Figure \ref{fig:one_shot_estimation}. It is interesting to see that for MNIST, our approximation is as effective as uncertainties estimated through iterative realizations of neural networks. For FMNIST (which is a more challenging dataset), AUE still produces respectable results. It is important to note that iterative procedure creates multiple instantiations of the network and every instance needs to see the entire dataset to estimate uncertainty. Although, AUE cannot match the performance of the iterative approach, it significantly reduces the time taken to calculate the uncertainty and this might be a worthy trade-off between performance and speed.
\section{Conclusion}

In this paper, we analyzed Bayesian neural networks for active learning and showed that they are more efficient than ensemble and Monte Carlo dropouts in capturing uncertainty. Our experiments revealed the following characteristics of BNN:
\begin{inparaenum}[(a)]
	\item BNNs are more robust when it comes to continual learning, and EN performs worse than MCD in this setting,
	\item produces respectable uncertainty estimates even with reduced model capacity, and
	\item produces better precision and F1 on difficult class labels compared to EN.
\end{inparaenum}
To overcome the computation cost of repeated uncertainty estimation in active learning, we proposed an accelerated uncertainty estimation technique for dense layers.
%From these experiments we can make ..ensembles are not always the better choice when it comes to active learning. One major draw back of ensemble is the size of parameters
%Even when retraining from scratch the performance of BNN is as good or in some cases better than EN and this is a significant ..Although BNN has twice the number of parameters when compared to MCD, in our setting EN has five times the number of parameters as MCD. 
%Sequential training of ENs takes a considerable time especially when retraining from scratch and training BNN is much faster. While ENs can be trained in a parallel fashion and can be faster than BNNs it comes at a cost of resource utilization. Additionally, there are recent works that propose various techniques to parallelize Bayesian training, which greatly speeds up the training time.\cite{beluch2018power} shows that variation-ratio is the best choice for acquisition function. However, we found that this is not the case for all scenario. 

{\small
\bibliographystyle{unsrt}
\bibliography{references}
}
\newpage
%\thispagestyle{empty}
%\mbox{}
%\newpage
%\input{supplementary}
\end{document}